\documentclass[a4paper,fleqn]{cas-sc}

\usepackage[square,numbers,sort&compress,comma]{natbib}

\def\tsc#1{\csdef{#1}{\textsc{\lowercase{#1}}\xspace}}
\tsc{WGM}
\tsc{QE}
\tsc{EP}
\tsc{PMS}
\tsc{BEC}
\tsc{DE}

\usepackage{caption}
\usepackage{soul}
\usepackage{subcaption}
\usepackage{xcolor}
\usepackage{xspace}
\usepackage{todonotes}
\usepackage{float}
\numberwithin{equation}{section} 

\newcommand{\projectname}{HURRI-GAN\xspace}

\newcommand{\theprojectname}{HURRI-GAN}

\begin{document}
\let\WriteBookmarks\relax
\def\floatpagepagefraction{1}
\def\textpagefraction{.001}

\shorttitle{HURRI-GAN}

\shortauthors{N. Nader et~al.}

\title [mode = title]{HURRI-GAN: A Novel Approach for Hurricane Bias-Correction Beyond Gauge Stations using Generative Adversarial Networks}                      



%
\author[1]{Noujoud Nader}[
                        orcid=0009-0000-4687-1416]

\cormark[1]


\ead{nnader@lsu.edu}


\credit{Methodology, Software, Validation, Formal analysis, Investigation, Visualization, Writing - Original Draft}

\affiliation[1]{organization={Center of Computation and Technology, Louisiana State University},
    city={Baton Rouge},
    postcode={70803 LA}, 
    country={US}}

\author[2]{Hadi Majed}[style=chinese]
\ead{hadi.majed@net.usj.edu.lb}
\credit{Methodology, Software, Validation, Formal analysis, Investigation, Visualization}
\affiliation[2]{organization={Saint-Joseph University of Beirut},
    city={Beirut},
    postcode={1104}, 
    country={Lebanon}}

\author[3,4]{Stefanos Giaremis}[%
   orcid=0000-0002-0107-3127
   ]
\ead{sgiaremi@physics.auth.gr}
\affiliation[3]{organization={Department of Physics, Aristotle University of Thessaloniki},
    city={Thessaloniki },
    postcode={54124}, 
    country={Greece}}
\affiliation[4]{organization={Center for Interdisciplinary Research and Innovation, Aristotle University of Thessaloniki},
    city={Thessaloniki},
    postcode={57001}, 
    country={Greece}}

\affiliation[5]{organization={Department of Computer Science, Louisiana State University},
    city={Baton Rouge},
    postcode={70803 LA}, 
    country={US}}

\affiliation[6]{organization={Oden Institute for Computational Engineering and Sciences, The University of Texas at Austin},
            city={Austin},
          citysep={}, 
            postcode={78712}, 
            state={TX},
            country={USA}}    
\credit{Methodology, Formal analysis, Investigation, Writing - Original Draft}

\author%
[2]
{Rola El Osta}
\ead{rola.osta@usj.edu.lb}
\credit{Formal analysis, Investigation}

\author[6]{Clint Dawson}[orcid=0000-0001-7273-0684]
\ead{clint@oden.utexas.edu}
\credit{Project administration, Funding acquisition}

\author%
[1]
{Carola Kaiser}
\ead{ckaiser@cct.lsu.edu}
\credit{Visualization, Supervision, Data Curation}

\author%
[1,5]
{Hartmut Kaiser} [orcid=0000-0002-8712-2806]
\ead{hkaiser@cct.lsu.edu}
\credit{Conceptualization, Supervision}




\begin{abstract}
The coastal regions of the eastern and southern United States are impacted by severe storm events, leading to significant loss of life and properties. Accurately forecasting storm surge and wind impacts from hurricanes is essential for mitigating some of the impacts, e.g., timely preparation of evacuations and other countermeasures. Physical simulation models like the ADCIRC hydrodynamics model, which run on high-performance computing resources, are sophisticated tools that produce increasingly accurate forecasts as the resolution of the computational meshes improves. However, a major drawback of these models is the significant time required to generate results at very high resolutions, which may not meet the near real-time demands of emergency responders.
The presented work introduces \theprojectname, a novel AI-driven approach that augments the results produced by physical simulation models using time series generative adversarial  networks (TimeGAN) to compensate for systemic errors of the physical model, thus reducing the necessary mesh size and runtime without loss in forecasting accuracy. We present first results in extrapolating model bias corrections for the spatial regions beyond the positions of the water level gauge stations.
The presented results show that our methodology can accurately generate bias corrections at target locations spatially beyond gauge stations locations. The model’s performance, as indicated by low root mean squared error (RMSE) values, highlights its capability to generate accurate extrapolated data. Applying the corrections generated by \projectname on the ADCIRC modeled water levels resulted in improving the overall prediction on the majority of the testing gauge stations. Such a model can act in conjunction with a temporal prediction approach as a component for real-time full spatiotemporal bias corrections to a physics-based model for operational forecasting systems. Moreover, it can potentially reduce the required resolution of the applied computational meshes without losing accuracy while also providing useful insight regarding the behavior of biases for future developments in storm surge modeling.
The source code is available on GitHub\footnote{\url{https://github.com/NoujoudNader/Extrapolation_GAN}} or Zenodo\footnote{\url{https://doi.org/10.5281/zenodo.15634528}}, respectively.


\end{abstract}




\maketitle

\section{Introduction}

    
Tropical cyclones are extreme weather events that affect coastal communities around the globe. In the United States alone, the associated annual damages between 1980 and 2024 have been estimated to exceed \$31 billion with 154 deaths per year on average, with these numbers being almost doubled in the last 20 years \cite{NCEI}. Moreover, severe storms, tropical cyclones and flooding constitute the top three natural disasters in descending order of occurrence frequency among all the billion-dollar disaster events that have taken place in the United States from 1980 to 2023, with two of these event types leading the list in both the total number of human casualties and total financial losses per year~\cite{NCEI2023}. Rise in sea surface temperature and other projected climate changes are predicted to increase the frequency of intense hurricanes and the magnitude of storm surge \cite{Camelo2020, Salarieh2023}. Therefore, the need for continuous improvement of numerical storm surge prediction frameworks in the context of operational warning systems is ongoing \cite{Bernier2024}.

Storm surge numerical models are typically based on the shallow-water equations in barotropic, depth-integrated form, with forcing from wind and atmospheric pressure, bottom drag, tides, and wind waves \cite{Kolar1994, Resio2008}. The ADvanced CIRCulation model for oceanic, coastal and estuarine waters (ADCIRC) 
is a high-fidelity hydrodynamic model that solves the shallow-water equations within the continuous Galerkin, linear finite element method on unstructured meshes, \cite{Westerink1992}. 
This approach has been 
extensively used for storm surge modeling \cite{Dietrich2011, Westerink2008, Hope2013}. ADCIRC is the main physics-based workhorse in many real-time forecasting frameworks, such as Coastal Emergency Risk Assessment (CERA). CERA is an interactive web visualization platform combining measurements from sources such as water level gauge stations and tide, wind-wave and hurricane storm surge numerical predictions, designed to provide first responders, decision makers and the general public with critical insights during hurricanes and extreme weather events \cite{CERA}. The CERA framework has also been recently used for the development of a historical storm archive containing hindcasts of more than 60 storms over the last 20 years \cite{CERA2023, CERAarchive}.

Recent improvements in storm surge and ocean circulation modeling in terms of mesh design and treatment of the description of natural processes have significantly improved accuracy and computational efficiency \cite{Pringle2021, Khani2023, Blakely2022, Loveland2024}. However, inherent 
uncertainties, although minimized, are inevitably present and well documented \cite{Gonzalez2019}. 
These can be due to inaccuracies in the description of hurricane characteristics such as track and wind speed and/or inputs such as description of coastal elements and land cover specification \cite{Ferreira2014, Torres2019, Gallien2018, Munoz2022}. Another source of uncertainty can be due to unresolved drivers such as rainfall, large-scale oceanic motions, and hydrological input, which are often not explicitly treated in storm surge models to restrict the complexity and computational cost of the model \cite{Asher2019}. Neglecting uncertainties has been shown to lead to biases that have a substantial impact on storm surge predictions and risk assessment \cite{Resio2012}. Therefore, detecting and quantifying these uncertainties is essential for improving the reliability of storm surge forecasting.

Traditional approaches for treating uncertainties and biases involve ensemble forecasting, 
data assimilation 
and other statistical methods such as quantile mapping \cite{Munoz2022, Asher2019, Butler2012, Hollt2015, Li2019}. More recently, machine learning (ML) approaches have been also explored for this purpose, showing improved accuracy in comparison with previous state-of-the art statistical methods \cite{Wang2022}. Commonly used ML architectures involve deep convolutional neural networks (CNNs), long-short term memory (LSTM) networks, bagged regression trees and multilayer perceptrons (MLPs) \cite{Wang2022, Giaremis2024, Sun2022, Campos2020, Ellenson2020}. Generative adversarial networks (GANs) constitute an emerging and highly promising type of neural networks for inferring the probability distribution that a given training set is drawn from, based on game theory -- in addition to traditional optimization techniques \cite{goodfellow2020generative, Goodfellow2020}. Despite originally implemented mostly for 2D/3D image reconstruction applications, GAN-based models have been recently shown promising performance for spatially bias correcting temperature, precipitation and wind predictions from climate models based on observed data \cite{Li2024, Franois2021, OSTA24, Pan2021}. 

\textbf{Motivation.} Previously, our group has demonstrated the viability of using LSTM-based models for predicting the offsets between observed and simulated water level values in gauge stations based on their past values, trained on historical storm data from the CERA platform \cite{Giaremis2024}. In this work, we propose a novel approach, \theprojectname, based on TimeGAN \cite{yoon2019time}, an extension of the original GAN approach for treating time-sequence data, to learn the correspondence between the aforementioned offset time series and geographic coordinates, so it can generate the temporal behavior of the former at arbitrarily given coordinates. In this way, offset time series at gauge stations could be extrapolated to any desired mesh point. 
To our knowledge, this is the first report of applying the TimeGAN approach for bias correcting water level data via spatiotemporal extrapolation. These results aim towards the development of improved bias correction components in real-time forecasting frameworks.\\

We summarize the main contributions of this paper as follows:
\begin{itemize}
    \item We propose a novel approach for bias correction based on generative artificial intelligence (GenAI), called \projectname, designed for spatiotemporal extrapolation of water level offsets.

    \item We introduce a new application of TimeGAN for hurricane-induced storm surge bias correction, where the model learns the mapping between sequential offset time series and geographic coordinates -- enabling offset generation at unseen locations on the mesh.

    \item We leverage historical storm data from the CERA platform for both model training and validation.

    \item We demonstrate the feasibility of AI-enhanced storm surge bias correction within real-time forecasting frameworks, especially for coastal regions beyond gauge stations aiming to improve the accuracy, reliability, and spatial coverage of storm surge predictions.
\end{itemize}
\section{Data and Methodology}\label{sec:data_and_methodology}

\begin{figure}[!ht]
	\centering
    		\includegraphics[width=1\textwidth]{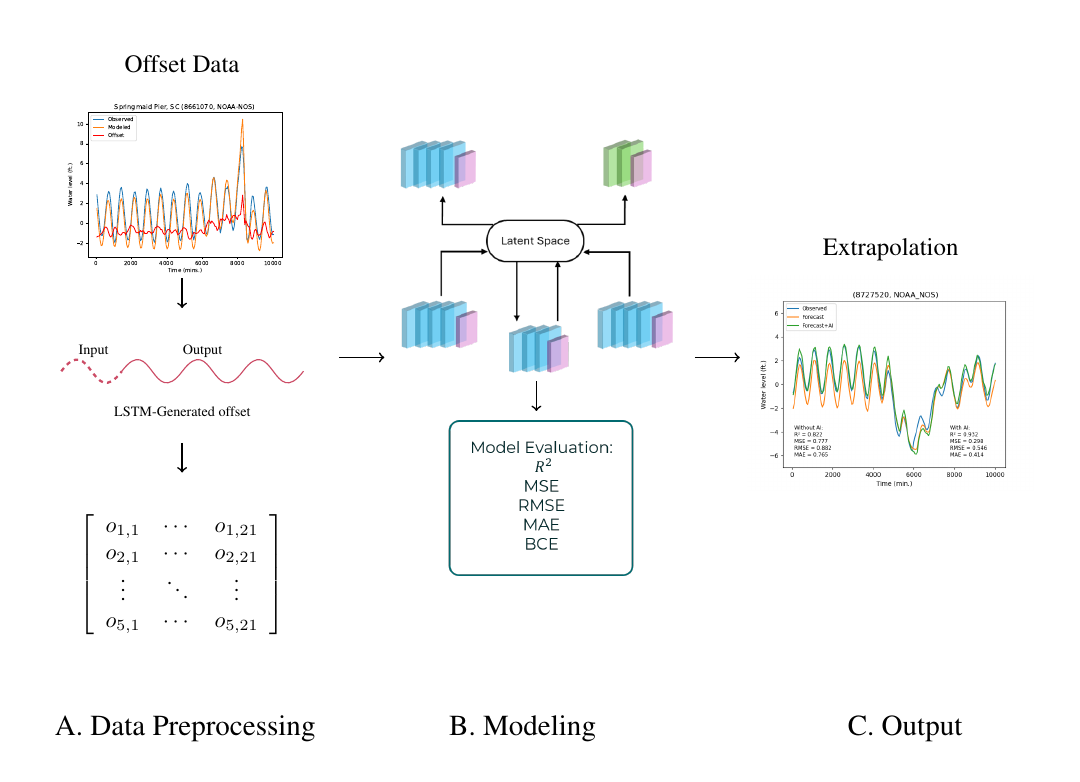}
	  \caption{Overview of the methodology framework: (A) Data Pre-processing phase includes offset extraction using Eq. \ref{eq:offsets}, data cleaning, and normalization. (B) Modeling phase involves the structure of the TimeGAN components and model evaluation using regression metrics. (C) Output phase includes the final application of the pre-trained model for bias extrapolation. The generated offsets are then used to correct the forecasted data using Eq. \ref{eq:corrected}.}
   \label{fig:pipeline}
\end{figure}
\subsection{Overview}
Figure~\ref{fig:pipeline} illustrates the methodology framework for \projectname proposed in this paper. It comprises three key steps, namely data pre-processing, modeling, and extrapolation. The first stage (Figure~\ref{fig:pipeline}.A) involves the systematic extraction of the offsets (Eq. \ref{eq:offsets}) between the modeled and observed water elevation time series from each gauge station in the available dataset. The offset time series for each gauge station is defined as follows: 

\begin{align}
    H_\text{offset}(t) = H_\text{modeled}(t) - H_\text{observed}(t) 
\end{align} \label{eq:offsets}
where $H_\text{modeled}(t)$ and $H_\text{observed}(t)$ are the forecast (via ADCIRC) and observed (from gauge stations) water levels, respectively, and $H_\text{offset}(t)$ is the water level offset (i.e., the bias), at each timestep, $t$. More details on this phase are explained in Section \ref{ssec:data}.
 In this work, we demonstrate the application of our newly proposed \projectname model to extrapolate previous ML-predicted biases to arbitrary spatial coordinates. We use our previous LSTM-based model \cite{Giaremis2024} to forecast the complete offset signals at the gauge stations, which, in turn are used to train our newly proposed GenAI model after appropriate reshaping and preprocessing (Section \ref{sssec:preprocessing}). The
GenAI model built to address our extrapolation problem is based on the TimeGAN approach \cite{yoon2019time} and constitutes the \projectname model. The processed samples are passed to the \projectname model for training and testing to assess its extrapolation performance. The gauge stations are divided into training and testing subsets. Testing is performed on the the testing gauge stations, which are not included by any means in the training of the model (see Section~\ref{sssec:clustering}), so that corrected water level forecasts can be directly evaluated against their observed counterparts. The structure and parameters of the model are optimized during this phase (Figure~\ref{fig:pipeline}.B). For more details, we refer to Section \ref{sec:genAI}. Once trained, the model can be applied for spatial extrapolation by giving as input only the coordinates of an arbitrary mesh point. After generating the offsets $H_\text{generated offsets}(t)$, the corrected forecast water level data, $H_\text{corrected}(t)$, are calculated according to the following equation:
\begin{align} \label{eq:corrected}
H_\text{corrected}(t) = H_\text{modeled}(t) - H_\text{generated offset}(t) 
\end{align} .
The model's extrapolation performance is evaluated based on the corrected water levels (Eq. \ref{eq:corrected}) for the selected testing gauge stations (Figure~\ref{fig:pipeline}.C).


\subsection{Data}
\label{ssec:data}
For this analysis, we use water level data for six hurricanes, obtained from the Historical Storm Surge Archive by \cite{CERA2023} and visualized through the built-in interface with the CERA website  \cite{CERA}. The hurricane data includes both modeled and observed water level values. The modeled values are produced by ADCIRC \cite{Westerink1992}. Observed data are obtained from different agencies including the National Oceanic and Atmospheric Administration (NOAA) \cite{NOAA}, coastal gauge stations (USGS) \cite{USGS}, U.S. Army Corps of Engineers (USACE) \cite{armyRivergagescomProviding}, Texas Coastal Ocean Observation Network (TCOON) \cite{noaaNOAATides}, and Puerto Rico Seismic Network (PRSN) \cite{uprmPuertoRico}. Both data sets are collected at hourly intervals. Offsets, as defined by Eq. \ref{eq:offsets}, are calculated by taking the difference between observed and modeled water level at each hourly time interval. This process resulted in the creation of an offset time series for each gauge station available during each hurricane. 
Gauge stations with missing offset values are excluded from the analysis. Additionally, an extra filtering step is performed to eliminate the influence of gauge stations that exhibited abnormal offset values. Stations with identified offset outlier values are also removed from this analysis. 
Table~\ref{tab:hurricanes_table} presents the hurricanes used in this study, along with the corresponding number of stations and the total amount of hourly offsets data collected for each hurricane after preprocessing.

\begin{table}[]
    \caption{Hurricanes considered in this study, their category based on the Saffir-Simpson hurricane scale~\cite{SaffirSimpson}, the number of station and the total amount of hourly offset data in each.}
        \label{tab:hurricanes_table}

    \centering
    \begin{tabular}{l|c|c|c}
       Hurricane  &Category & No. of stations & No. of hourly offsets 
       \\ \toprule
       Ian (2022)  & H5 & 250 & 26250 \\
        Harvey (2017)  & H4 & 247 & 25935 \\
       Ida (2021)  & H4 & 264 & 18216 \\
       Idalia (2023)  & H4 & 304 & 31920 \\
       Matthew (2016)  & H4 & 236 & 24780 \\
       Hermine (2016)  & H1 & 259 & 27195 \\

    \end{tabular}
\end{table}
\subsubsection{Clustering}
\label{sssec:clustering}
\begin{figure}
    \centering
    \includegraphics[width=0.5\linewidth]{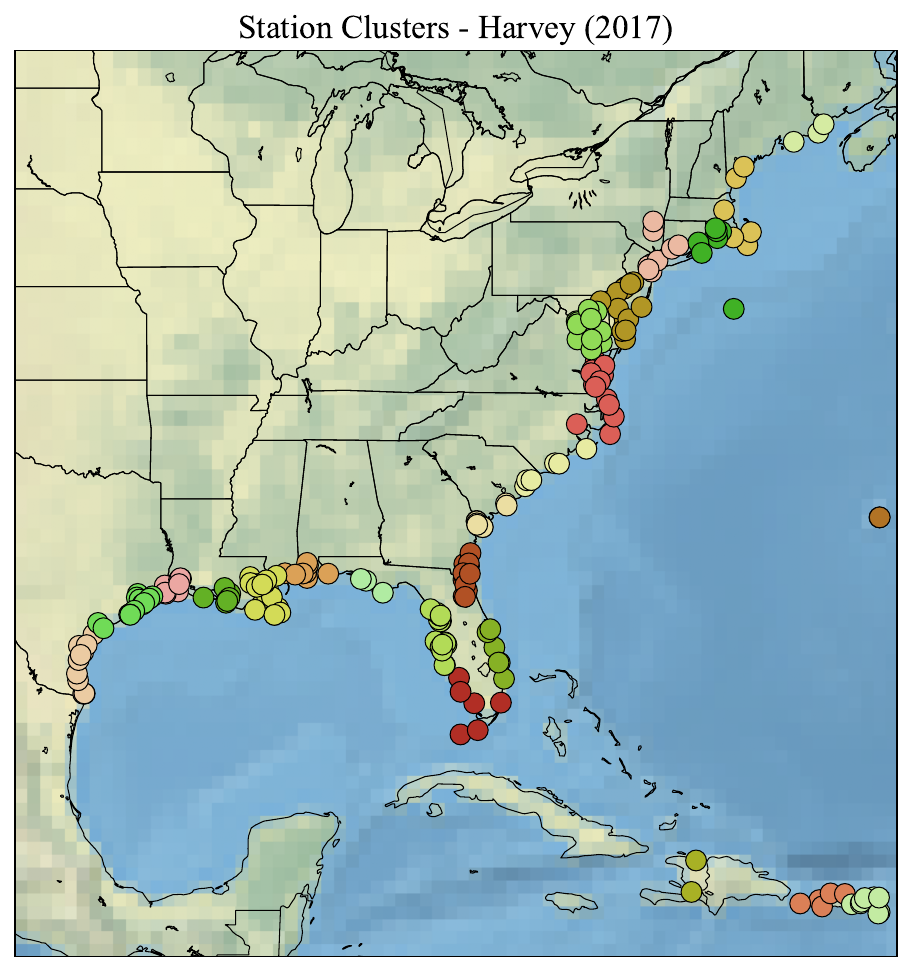}
    \caption{Station clustering for Hurricane Harvey (2017). This map illustrates the clustering of stations based on their coordinates. A total of 24 clusters were identified, with each color representing a different cluster.}

        \label{fig:harvey_clusters}
\end{figure}

The next step is to split the gauge stations within each hurricane into training and testing sets.
As we are working with a GenAI model, we need to provide more training data than usual, as the task requires more data for proper learning. For this, a split of 90\% of the stations set for training and 10\% for testing is mainly chosen. To ensure that each region in the studied geographic area is properly represented, we employ the K-means clustering algorithm \cite{sinaga2020unsupervised} to geographically divide the stations into groups. The gauge stations are clustered into groups based on their coordinates, with the condition that each cluster contains more than one station. For each hurricane, the number of clusters is 10\% of the number of stations. This clustering approach helps to ensure that the testing stations are not concentrated in a single region but are representative of the entire geographic study area. 
From each cluster, one station is randomly selected and designated as a testing station while the remaining stations are added to the training set. An example of the clustering process for Hurricane Harvey(2017) is illustrated in Figure~\ref{fig:harvey_clusters}, where the stations are divided into 24 clusters based on their coordinates, ensuring a proper representation of all geographical areas in the testing set. Each cluster group is presented with a different color.

\subsubsection{Preprocessing}
\label{sssec:preprocessing}
After splitting the gauge stations into training and testing sets, the offsets are normalized for each hurricane. \texttt{MinMax} scaling from Scikit-Learn \cite{pedregosa2011scikit} is used on the training and testing data separately to avoid data leakage. The time series data for each gauge station is then reshaped into a 2D array format to be compatible with the input requirements of the model \cite{yoon2019time}. In this case, data from each gauge station are fed into the model as a unique input, with its offset data reshaped into a (5 x 21) array for most hurricanes, where the total signal length is 105. For Hurricane Idalia, which has a signal length of 69 hours, the data is reshaped into a (3 x 23) array. The data is then distributed into batches, with the batch size set to 10. The choice of batch size ensures efficient model training while managing memory consumption.
As for the coordinates, four static embeddings were retrieved from each pair and are repeated for each row in the input matrix. For each timestep in this matrix, the model can properly relate the coordinate’s embeddings to the temporal sequence. Having a large number of embeddings will make it too complicated for the model to learn, while having different embeddings for each timestep will result in the model failing to capture consistent spatial understanding.
Finally, each batch contains the gauge station's coordinate tuple (x, y) paired with its corresponding offset matrix. The input shape for each sample is as follows: 
\begin{equation*}
 \left[
\begin{array}{cccccccccccccc}
o_{1,1} & \cdots & o_{1,21} \\
o_{2,1} & \cdots & o_{2,21} \\
\vdots & \ddots & \vdots \\
o_{5,1} & \cdots & o_{5,21}
\end{array}
\right]
\left[
\begin{array}{cccc}
y_1 & \cdots & y_4 \\
y_1 & \cdots & y_4 \\
\vdots & \ddots & \vdots \\
y_1 & \cdots & y_4
\end{array}
\right]
\end{equation*}


\subsection{GenAI models}
\label{sec:genAI}
Generative artificial intelligence (GenAI) models represent a groundbreaking stride in the realm of artificial intelligence, facilitating the creation of novel data rather than solely interpreting existing information. Notable examples such as GANs and sophisticated autoregressive models like the distinguished OpenAI GPT series excel in generating lifelike images, text, music, and diverse content forms. By discerning intricate patterns from extensive datasets, these generative AI models exhibit the capacity to craft outputs that often mirror human-created content. This innovation carries profound implications spanning various domains, including art, design, content generation, and its potential utility in accelerating drug discovery and scientific exploration \cite{Akh24, Sen24}. In the following, we will discuss the overview of TimeGAN components, and the proposed \projectname structure.

\subsubsection{Overview of TimeGAN Components}
To address the challenge of generating data that matches the temporal distribution of a real data sample, Yoon and Jarrett \cite{yoon2019time} proposed a new GAN architecture called TimeGAN. In traditional GANs \cite{goodfellow2020generative}, a neural network, referred to as the generator (\texttt{G}), aims to map random noise to a target distribution. An auxiliary neural network, known as the discriminator (\texttt{D}), guides the training of \texttt{G} by distinguishing between the generated data and the real data (i.e. offsets). This process, known as adversarial learning, involves \texttt{G} being trained not toward a fixed objective, but rather to fool \texttt{D}, which is concurrently updated to improve its discrimination capabilities.

TimeGAN \cite{yoon2019time} introduces three additional components with corresponding loss functions. The embedder (\texttt{E}) and recovery (\texttt{R}) models together form what is known as an autoencoder. The embedder's role is to map real data samples, which incorporate both temporal and static features, into an abstract representation known as the latent space. This space contains latent codes or embeddings. Specifically, the embedder maps both temporal features (time series offsets) and static features (gauge station coordinates) into this latent space, while the recovery model reconstructs these embeddings back into their original representations. As the feedback from the discriminator \texttt{D} may not be enough, a new component, the supervisor (\texttt{S}), is introduced to bridge the gap between the autoencoder and adversarial networks. This supervisor helps the embedder generate better embeddings, which in turn improves the learning process for the generator. 

\subsubsection{Proposed \projectname structure}
A detailed structure of the network for each temporal component used in \projectname is presented in Table \ref{tab:gan_structure}.
The embedder component, as the generator and recovery, consists of two neural networks, one that deals with the temporal data (i.e. offsets time-series) and one with the static data (i.e. the stations coordinates). The static network is a simple dense layer with four neurons and ‘\textit{sigmoid}’ activation function that maps the coordinate components into four latent representations or two reconstructed coordinates for the recovery. The temporal network is a much more complex model with multiple Gated Recurrent Unit (GRUs) layers followed by a dense layer with a number of neurons equal to the number of columns in the input matrix (Section \ref{ssec:data}). The supervisor follows the same architecture but with a smaller number of GRU layers. The discriminator architecture consists of only a temporal network with a two bi-directional GRU layers, followed by a dense layer with one neuron corresponding to its binary classification task.
For all of the components containing static networks, the coordinate input passes first through the static network. The resulting output is concatenated with the temporal input, and fed into the temporal networks that will study the relationship between the two types of inputs and provide the results accordingly. The structure and the number of neurons were selected based on hyperparameter tuning (Section \ref{sec:hypertuning}).

\begin{table}[htbp]
\centering
\caption{Schematic overview of the structure of the \projectname model (GRU: Gate Recurring Unit, Bi-GRU: Bidirectional Gated Recurrent Unit).}
\begin{tabular}{llll}

\\
\multicolumn{2}{l}{\textbf{Embedder, Recovery, Generator}} & 

\\
\hline
 \textbf{Layers} & \textbf{No. of Neurons} & \textbf{Activation} & \textbf{Output Shape} \\ 
 \hline
GRU & 256 &  -- & 5*256\\
GRU & 256 &  -- & 5*256\\
GRU & 256 &  -- & 5*256\\
GRU & 256 &  -- & 5*256\\
GRU & 256 &  -- & 5*256\\
Dense & 21 & Sigmoid & 5*21\\

\hline
\\
\multicolumn{2}{l}{\textbf{Supervisor}} & 

\\
\hline
 \textbf{Layers} & \textbf{No. of Neurons} & \textbf{Activation} & \textbf{Output Shape} \\ 
 \hline
GRU & 256 &  -- & 5*256\\
GRU & 256 &  -- & 5*256\\
GRU & 256 &  -- & 5*256\\
GRU & 256 &  -- & 5*256\\
Dense & 21 & Sigmoid & 5*21\\
\hline

\\
\multicolumn{2}{l}{\textbf{Discriminator}} & 

\\
\hline
 \textbf{Layers} & \textbf{No. of Neurons} & \textbf{Activation} & \textbf{Output Shape} \\ 
 \hline
Bi-GRU & 256 &  -- & 5*512\\
Bi-GRU & 256 &  -- & 1*512\\
Dense & 1 & Sigmoid & 1*1\\
\end{tabular}
\label{tab:gan_structure}
\end{table}

To train the \projectname model, we first train the autoencoder (\texttt{E} and \texttt{R}). 
 The autoencoder learns from real data (offsets) by minimizing temporal and static embedding losses, both calculated using mean squared error (MSE, Eq.~\ref{eq:mse}). The autoencoder's parameters are updated based on the gradients to minimize the loss.
Alternatively, we train \texttt{S} using real embeddings from the embedder by  minimizing the MSE loss (Eq.~\ref{eq:mse}).
Finally, all the components are trained together. The generator and embedder are trained more frequently than the discriminator, which is only trained when its loss exceeds a threshold. The generator's loss is computed from a combination of adversarial, supervisor, and distribution losses, while the discriminator's loss is based on comparing values of Binary Cross-Entropy (BCE):
\begin{equation}
\text{BCE} = - \frac{1}{n} \sum_{i=1}^{n} [y_i \log(\hat{y}_i) + (1 - y_i) \log(1 - \hat{y}_i)]
\label{eq:bce}
\end{equation}
across real and generated data.  In Eq.~\ref{eq:bce}, $y$ represents the real label of the data and $\hat{y}$ represents the predicted label of the data. The Adam optimizer \cite{kingma2014adam} was used for all of the compiled models during training. 

\subsection{Evaluation Parameters and Computational Details}
\label{sec:eval}
Mean square Error (MSE, Eq.~\ref{eq:mse}), root mean squared error (RMSE,  Eq.~\ref{eq:rmse}) and mean absolute errors (MAE,  Eq.~\ref{eq:mae}) 
are used here as metrics to evaluate the model performance. Smaller values of errors 
mean higher generation accuracy.
\begin{equation}
\text{MSE} = \frac{1}{n} \sum_{i=1}^{n} (y_i - \hat{y}_i)^2
\label{eq:mse}
\end{equation}

\begin{equation}
\text{RMSE} = \sqrt{\frac{1}{n} \sum_{i=1}^{n} (y_i - \hat{y}_i)^2}.
\label{eq:rmse}
\end{equation}

\begin{equation}
\text{MAE} = \frac{1}{n} \sum_{i=1}^{n} \rvert y_i - \hat{y}_i \rvert
\label{eq:mae}
\end{equation}

To evaluate the general performance of the extrapolation model \projectname, we first evaluated MSE and MAE between real and generated offsets. In this case,  $y$ represents the real offset values, $\hat{y}$ stands for the extrapolated (predicted) offset values and $n$ denotes the total number of samples. Then we evaluated the bias correction after generation, these
evaluation metrics are estimated between the corrected forecast (Eq.~\ref{eq:corrected}) and the original forecast (produced by ADCIRC) . Herein, $y$ represents the original forecast values (without AI), $\hat{y}$ stands for the the corrected forecast (with AI).

The \projectname model is 
implemented using TensorFlow 2.16.2 with Keras in Python 3.9.18.

\section{Results and Discussion} 
\subsection{LSTM-based model results}
As a first step, offsets predictions at known gauge station locations are generated by using our previously developed LSTM-based model \cite{Giaremis2024}. To estimate the accuracy of the LSTM-based model, the evaluation metrics between real and LSTM-generated offsets for each hurricane, including MSE, RMSE, and MAE, are presented in Table \ref{tab:lstm_test}.  As shown in Table \ref{tab:lstm_test}, the LSTM-based model demonstrated strong performance across all storms, yielding consistently low MSE, RMSE, and MAE values. These results, which are in agreement with our previously published results, confirm the LSTM-based model’s reliability and suitability for generating offset time series to train \projectname.

\begin{table}[h!]
\caption{Metrics in feet of the real vs ML-predicted offsets generated via our previous LSTM-based model \cite{Giaremis2024} for each of the considered hurricanes.}

\label{tab:lstm_test}

\centering
\begin{tabular}{l|c|c|c}

\textbf{Hurricane} & \textbf{MSE} & \textbf{RMSE} & \textbf{MAE} \\
\toprule
 Ian (2022)& 0.129 & 0.359 & 0.241 \\
 Harvey (2017)& 0.121 & 0.347 & 0.214 \\ 

 Ida (2021)& 0.116 & 0.34 & 0.197 \\

 Idalia (2023)& 0.124 & 0.352 & 0.22  \\

 Matthew (2016)& 0.066 & 0.257 & 0.169 \\

 Hermine (2016)& 0.085 & 0.292 & 0.177  \\

\end{tabular}
\end{table}
\subsection{Hyperparameter tuning}
\label{sec:hypertuning}

To optimize our network architecture for each component in \projectname, we conduct hyperparameters tuning for each hurricane. In this process, stations are divided into training and testing sets. Subsequently, candidate models with different hyperparameters are trained on the former and tested on extrapolating data on the latter. Tuning is performed separately for each of the hurricane considered in this work. The parameters tuned during this process include the number of layers, number of neurons and number of epochs.  The number of layers is analysed for the generator, embedder and recovery. The supervisor is one layer less than the other componenets. The discriminator is two layers.  The architecture that gives the lowest RMSE in most cases is finally selected. 

 In our analysis (Table~\ref{tab:hpt_results}), it is observed that using 2000 training epochs for the joint training phase does not lead to the best performance for most hurricanes, suggesting that more training epochs are needed to adequately capture the complexity of the storm surge data. Moreover, the results from the hyperparameter tuning process reveal that 
 the optimal architecture varies in terms of layers, neurons, and epochs, highlighting the importance of conducting hurricane-specific hyperparameter optimization to ensure the best possible extrapolation performance within each hurricane. More specifically, for some hurricanes, a model with four layers provides the best performance (Hurricane Harvey (2017), Idalia (2023), Matthew (2016), and Hermine (2016)), while others require five layers to effectively capture the underlying data distribution (Table~\ref{tab:hpt_results}). Similarly, the optimal number of neurons and epochs varies depending on the complexity of the hurricane's data. However, the choice of 256 neurons and 3000 epochs is generally shown to yield optimal results for most storms (Table~\ref{tab:hpt_results}). Therefore, using the aforementioned settings, along with using 4 or 5 layers, would be expected to yield optimal results for a new storm in a real-world scenario. 
 \\

\begin{table}[tb]
\centering
\caption{Hyperparameter tuning results for the \projectname model. The table presents the RMSE values (in feet) along with the optimal hyperparameters for each hurricane, including the number of network layers, number of neurons, and number of epochs. The values for each of the aforementioned parameters considered in this analysis are listed in brackets. The number of layers is analyzed for Generator, Embedder and Recovery. The supervisor is one layer less than the other components, while the discriminator architecture always consists of two layers.  
}
\resizebox{1\textwidth}{!}{%
\begin{tabular}{c|c|c|c|c|c}
\cmidrule(lr){2-6}
   
\multirow{3}{*}{} & \multirow{3}{*}{\textbf{Error Metric}} & \multirow{3}{*}{\textbf{RMSE}} & \multicolumn{3}{c}{\textbf{Parameters}} \\

\cmidrule(lr){4-6}
& &  & \textbf{No. of Layers} & \textbf{No. of neurons} & \textbf{No. of Epochs} \\
 & & (ft.) & [4, 5] & [128, 256]  & [2000, 3000, 4000] \\
\midrule
\multirow{4}{*}{ \rotatebox{30}{\textbf{Ian(2022)}}} &Min        & 0.654  & 5  & 256    & 4000  \\
&Max        & 0.724& 4& 128&  4000\\
&Mean - Std & 0.669& 5&256&  3000\\
&Mean + Std & 0.707& 4& 256& 2000\\
\midrule
\multirow{4}{*}{ \rotatebox{30}{\textbf{Harvey(2017)}}} &Min        & 0.356  & 4  & 256    & 3000  \\
&Max        & 0.459& 5& 128&  2000\\
&Mean - Std & 0.36& 5&256&  3000\\
&Mean + Std & 0.415& 4& 128& 3000\\
\midrule
\multirow{4}{*}{ \rotatebox{30}{\textbf{Ida(2021)}}} &Min        & 0.623  & 5  & 128    & 3000  \\
&Max        & 0.809& 4& 128&  3000\\
&Mean - Std & 0.65& 4&256&  3000\\
&Mean + Std & 0.749& 5& 128& 4000\\
\midrule
\multirow{4}{*}{ \rotatebox{30}{\textbf{Idalia(2023)}}} &Min        & 0.362  & 4  & 256    & 4000  \\
&Max        & 0.523& 5& 128&  2000\\
&Mean - Std & 0.362& 4&256&  4000\\
&Mean + Std & 0.431& 5& 128& 4000\\
\midrule
\multirow{4}{*}{ \rotatebox{30}{\textbf{Matthew(2016)}}} &Min        & 0.275  & 4  & 128    & 3000  \\
&Max        & 0.465& 5& 128&  3000\\
&Mean - Std & 0.291& 4&256&  2000\\
&Mean + Std & 0.417& 4& 256& 3000\\
\midrule
\multirow{4}{*}{ \rotatebox{30}{\textbf{Hermine(2016)}}} &Min        & 0.368  & 4  & 128    & 3000  \\
&Max        & 0.477& 5& 256&  2000\\
&Mean - Std & 0.374& 4&256&  4000\\
&Mean + Std & 0.447& 4& 256& 2000\\
\bottomrule

\end{tabular}
}
\label{tab:hpt_results}
\end{table}


\subsection{Extrapolation analysis}
\begin{figure}[H]
    \centering
    \includegraphics[width=1\linewidth]{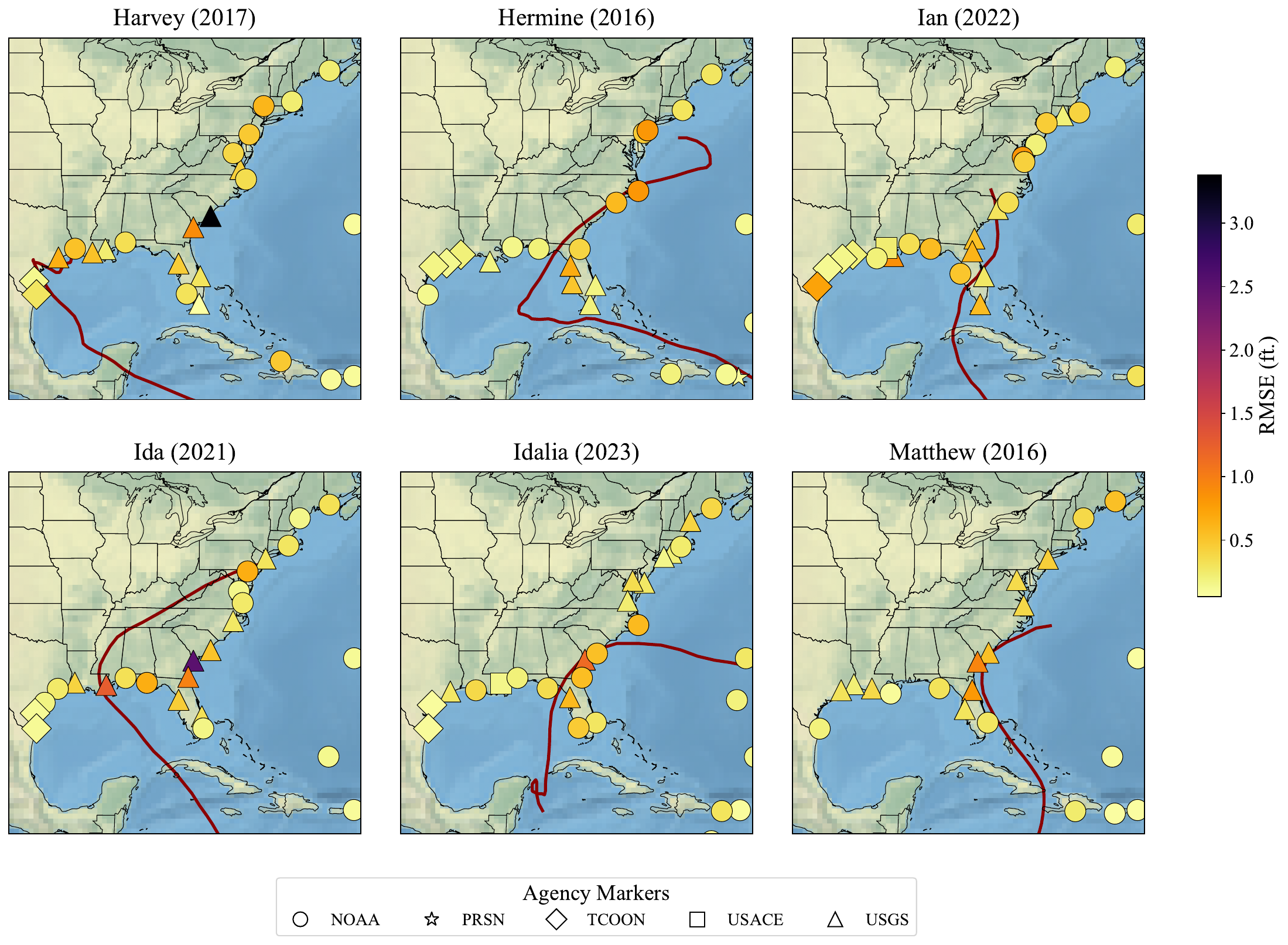}
    \caption{RMSE values for the testing stations across different hurricanes. The map for each hurricane shows the locations of the testing stations along with their corresponding RMSE values (in feet), indicated by the color gradient. The considered hurricanes are Harvey (2017), Hermine (2016), Ian (2022), Ida (2021), Idalia (2023), and Matthew (2016) and the considered agencies are NOAA, USACE, USGS, TCOON, and PRSN. The hurricane paths are outlined in red. 
    }
    \label{fig:stations_map}
\end{figure}


In this section, the behavior of the best model variation for each hurricane, based on the discussion in the previous section, will be analyzed in further detail.
From Table~\ref{tab:hpt_results}, it can be seen that, overall, the \projectname architecture demonstrates promising performance across all hurricanes, with the lowest RMSE (0.275 ft. or 0.083 m.) observed for Hurricane Matthew (2016), where the model effectively captures the storm surge dynamics with minimal error, thus indicating strong predictive accuracy for extrapolation. On the other hand, the model exhibits the highest RMSE (0.654 ft. or 0.199 m.) for Hurricane Ian (2022). 

To evaluate the performance of the model at individual locations, we analyze the RMSE of the extrapolated offsets generated by \projectname for each testing gauge station. 
Figure~\ref{fig:stations_map} presents the geographical distribution of the RMSE of the extrapolated offsets in each station for each hurricane, along with the corresponding station agency and the hurricane track~\cite{NOAA,CERAarchive}.


 Typically, RMSE values below 1.5 ft. (0.46 m) are observed in most cases. However, in some cases, stations located along the hurricane's path tend to have relatively higher RMSE values compared to those situated further inland or outside the path. This is expected due to the increased complexity 
 of storm surge dynamics in areas affected directly by the hurricane, particularly near landfall. For example, in the case of Hurricane Ida (2021), stations near the path of the hurricane (especially New Orleans at Louisiana) exhibit higher RMSE values. 
Moreover, a few outliers are observed, such as for Hurricanes Harvey (2017) and Ida (2021), in South Carolina and Georgia, respectively. Stations with higher RMSE values, not located directly in the path of the hurricane, are also observed in some other cases (such as for Hurricane Ian (2022) in a few stations in Texas, Louisiana and Virginia), although with not so pronounced RMSE values. Typically these cases correspond to stations located further inland or in areas with complex coastline features involving elements such as levies and jetties.  



\begin{figure}
    \centering
    \includegraphics[width=0.6\linewidth]{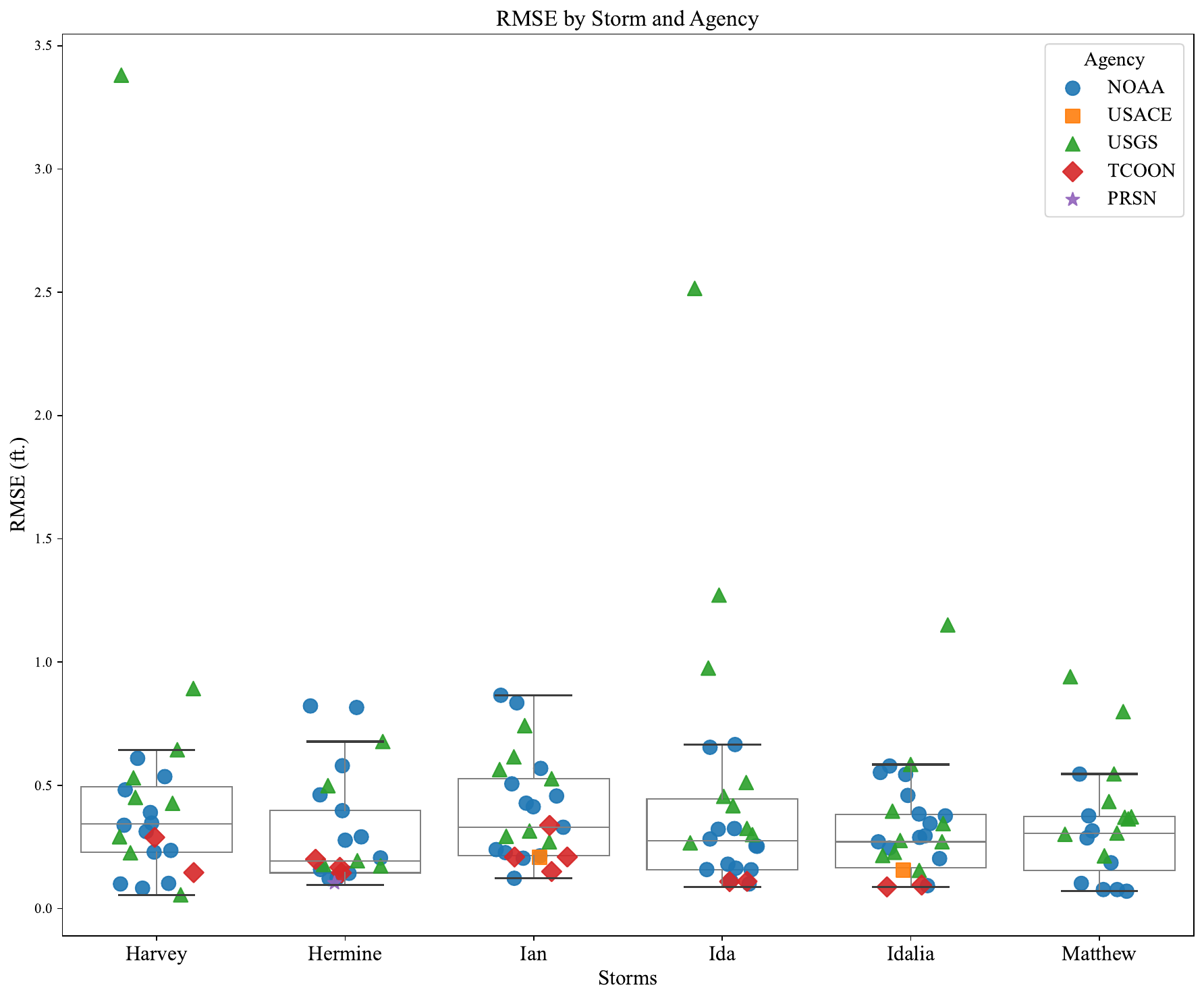}
    \caption{RMSE distribution of the extrapolated offsets with the \projectname model in the testing stations by hurricane and agency. The considered hurricanes are Harvey (2017), Hermine (2016), Ian (2022), Ida (2021), Idalia (2023), and Matthew (2016) and the considered agencies are NOAA, USACE, USGS, TCOON, and PRSN.}
    \label{fig:rmse_distribution}
\end{figure}

To further investigate this behavior, Figure~\ref{fig:rmse_distribution} provides a more detailed view of the distribution of RMSE  across different hurricanes and agencies. Each boxplot represents the RMSE distribution of testing stations for a given hurricane. The spread of values within each plot illustrates the variability in forecast accuracy across different locations. 
In the case of Hurricane Harvey (2017), higher RMSE values are typically found at stations located along the storm track, reflecting the greater complexity of storm surge dynamics near landfall. However, a common and consistent trend across all the considered hurricanes is that the highest RMSE values are predominantly observed at USGS stations, regardless whether the stations are in proximity to the storm track or not. USGS stations are often located further inland, e.g., along rivers, or in coastal areas with intricate coastline features, and thus governed by different hydrodynamical laws compared to stations more directly exposed to the ocean. 

\begin{figure}[tb!]
    \centering
    \begin{subfigure}[b]{0.25\textwidth}
            \centering
            \includegraphics[height=0.16\paperheight]{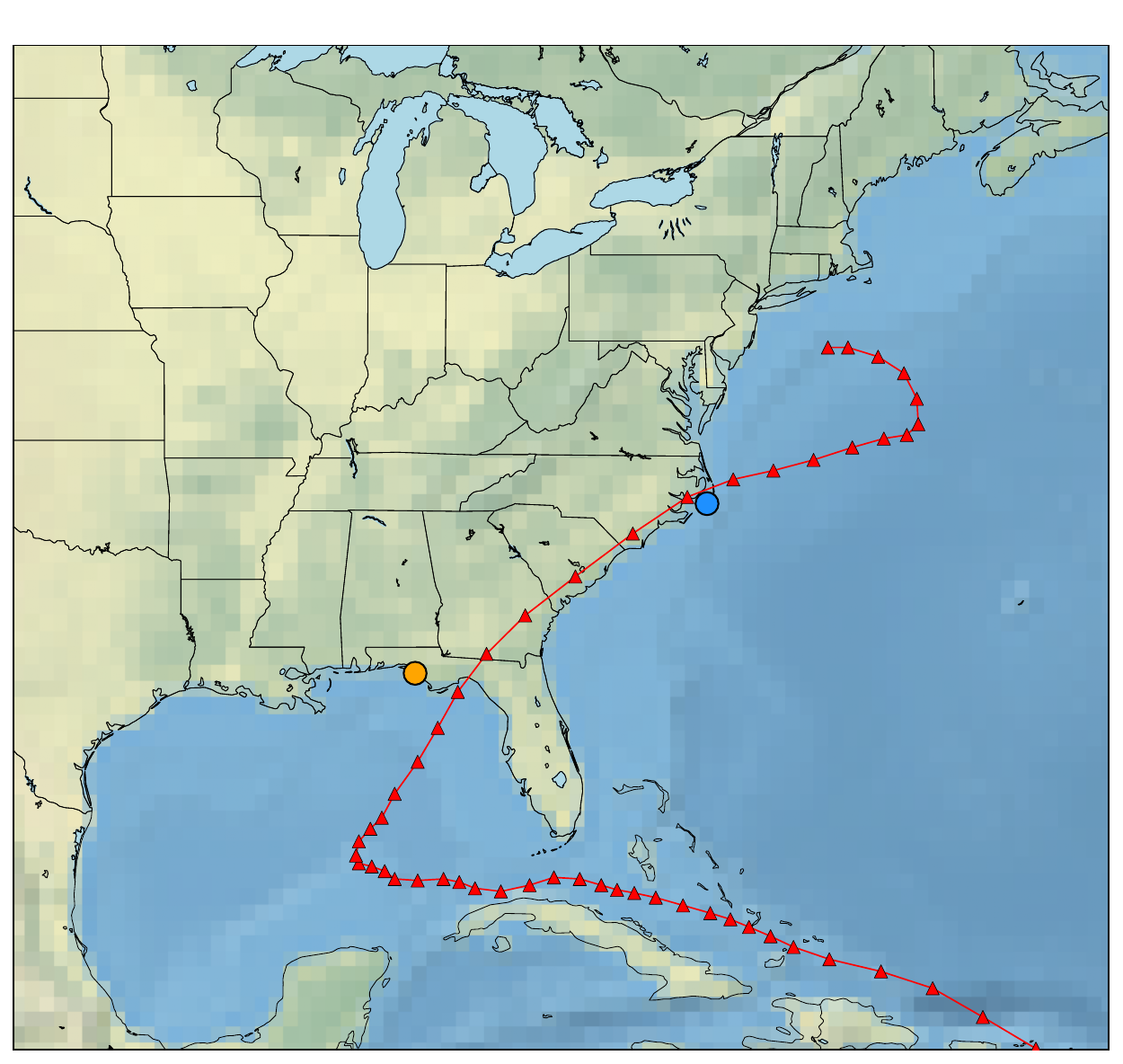} 
        \end{subfigure}
    \begin{subfigure}[b]{0.36\textwidth}
            \centering
            (a) Hurricane Hermine (2016)
            \includegraphics[width=\textwidth]{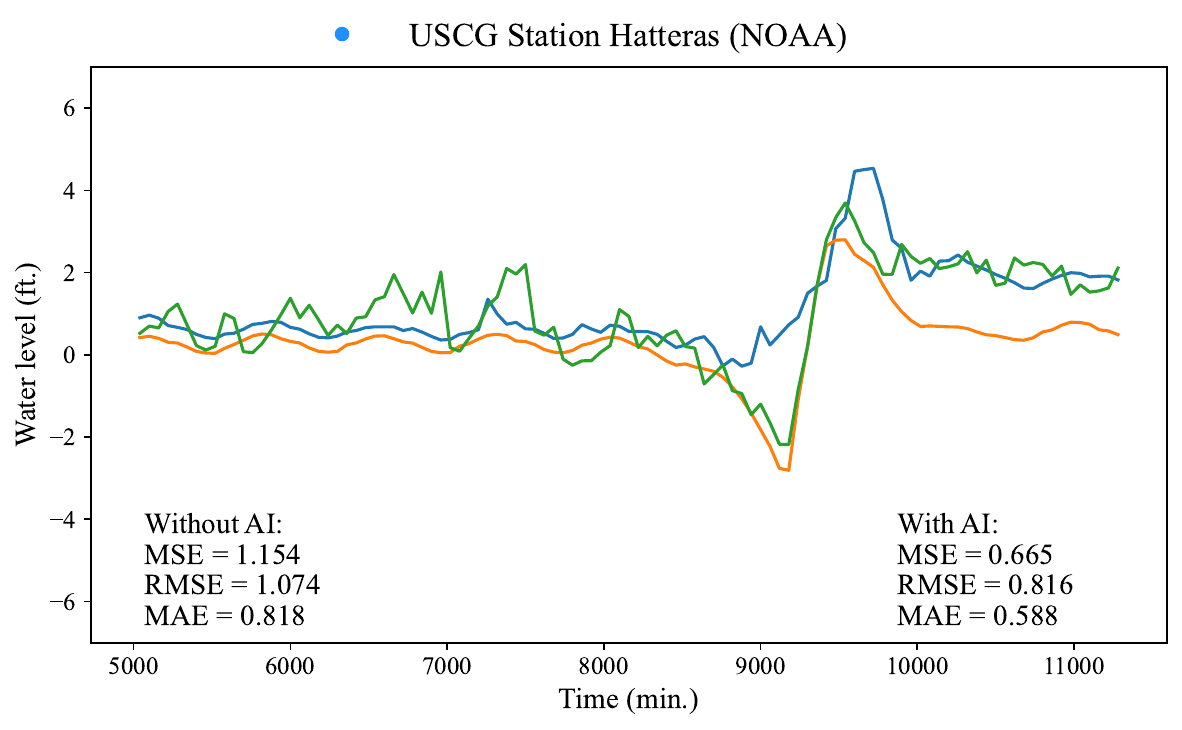}
        \end{subfigure}      
    \begin{subfigure}[b]{0.36\textwidth}
            \centering
            \includegraphics[width=\textwidth]{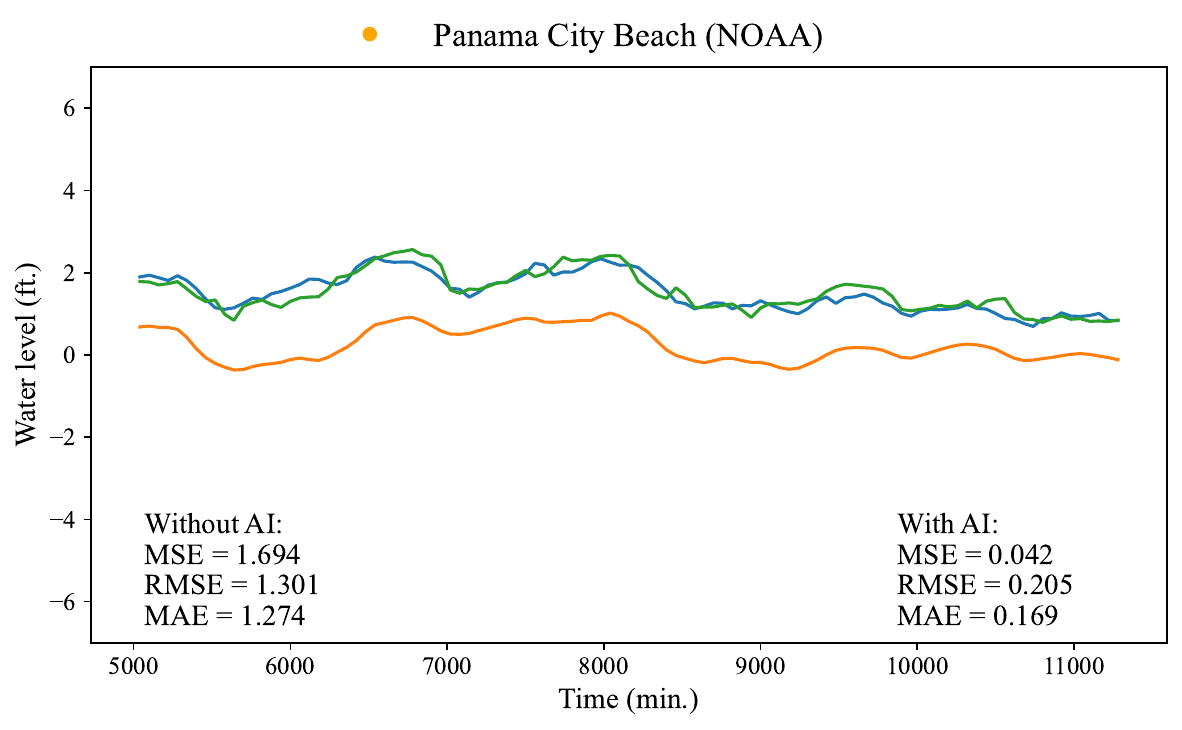}
        \end{subfigure}
    \begin{subfigure}[b]{0.25\textwidth}
            \centering
            \includegraphics[height=0.16\paperheight]{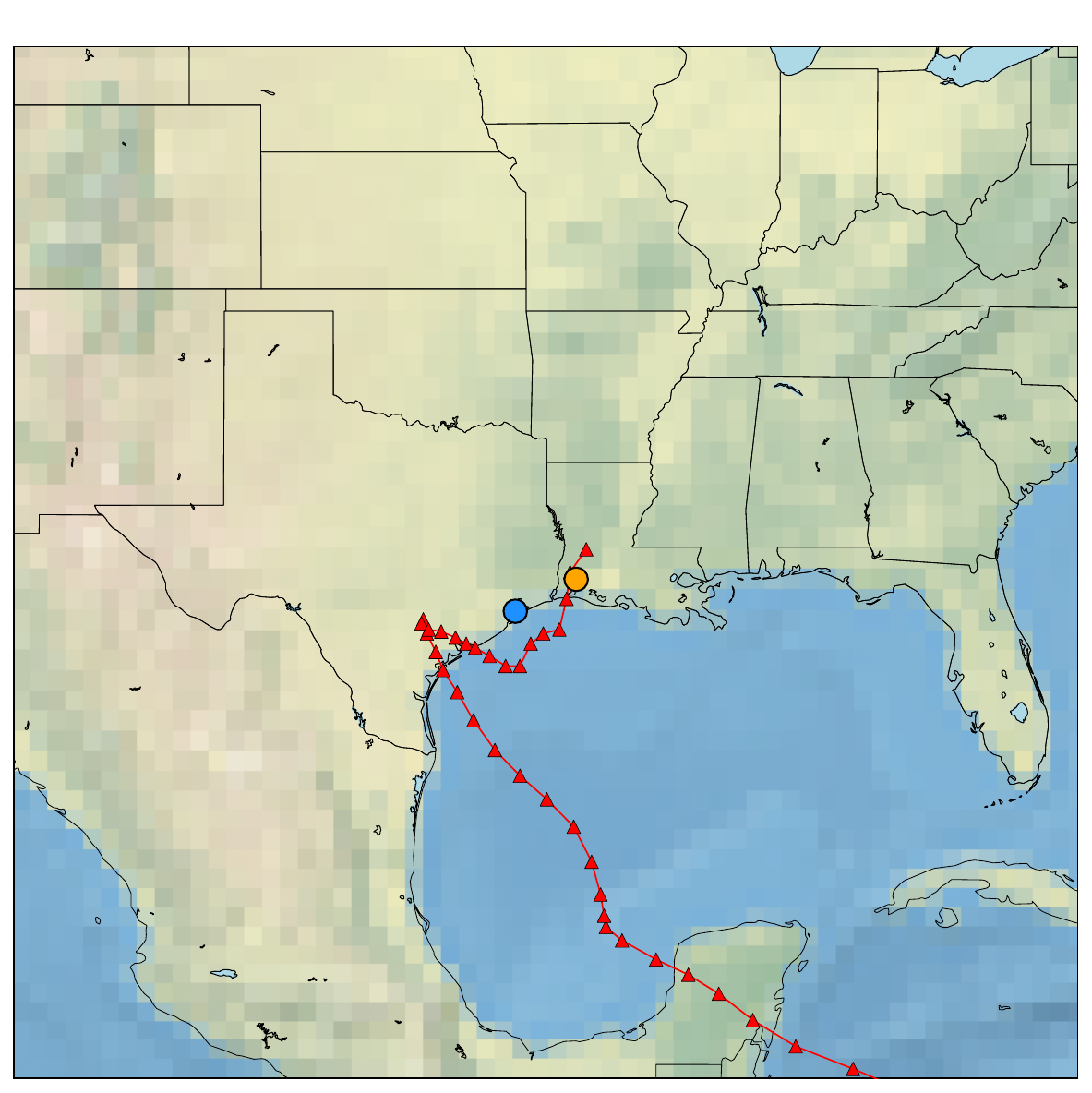}  
        \end{subfigure}
    \begin{subfigure}[b]{0.36\textwidth}
            \centering
            (b) Hurricane Harvey (2017)
            \includegraphics[width=\textwidth]{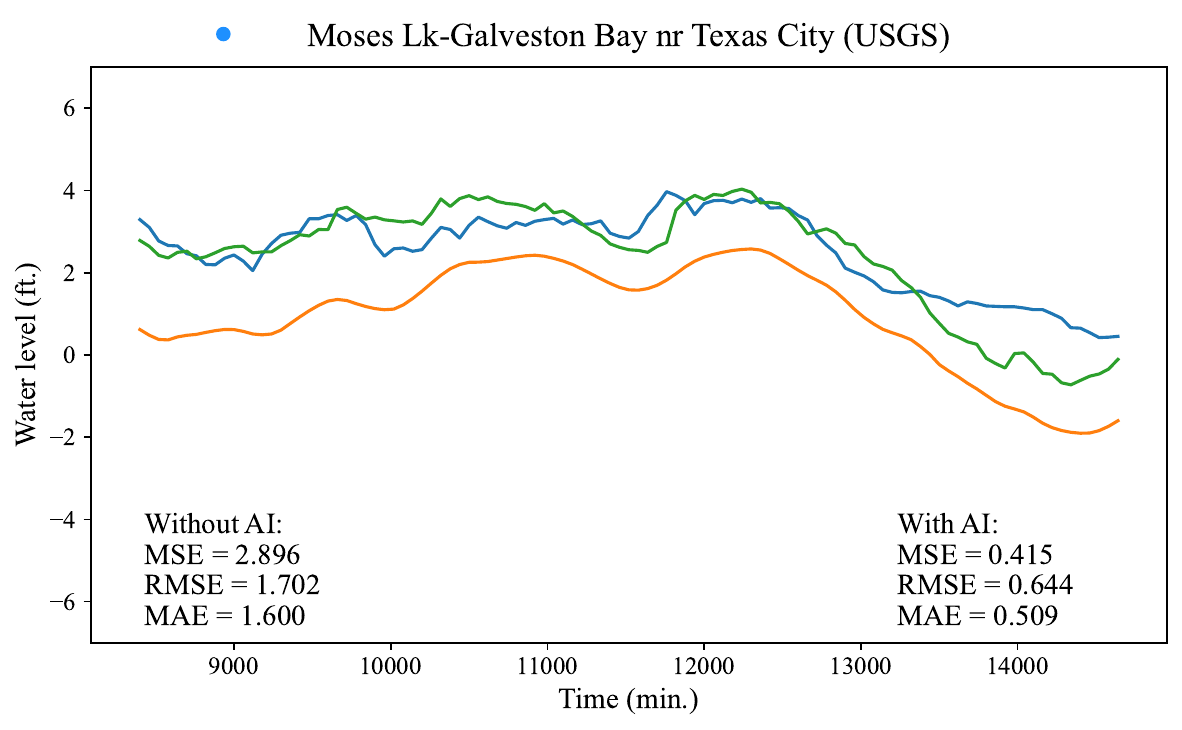}
        \end{subfigure}
    \begin{subfigure}[b]{0.36\textwidth}
            \centering
            \includegraphics[width=\textwidth]{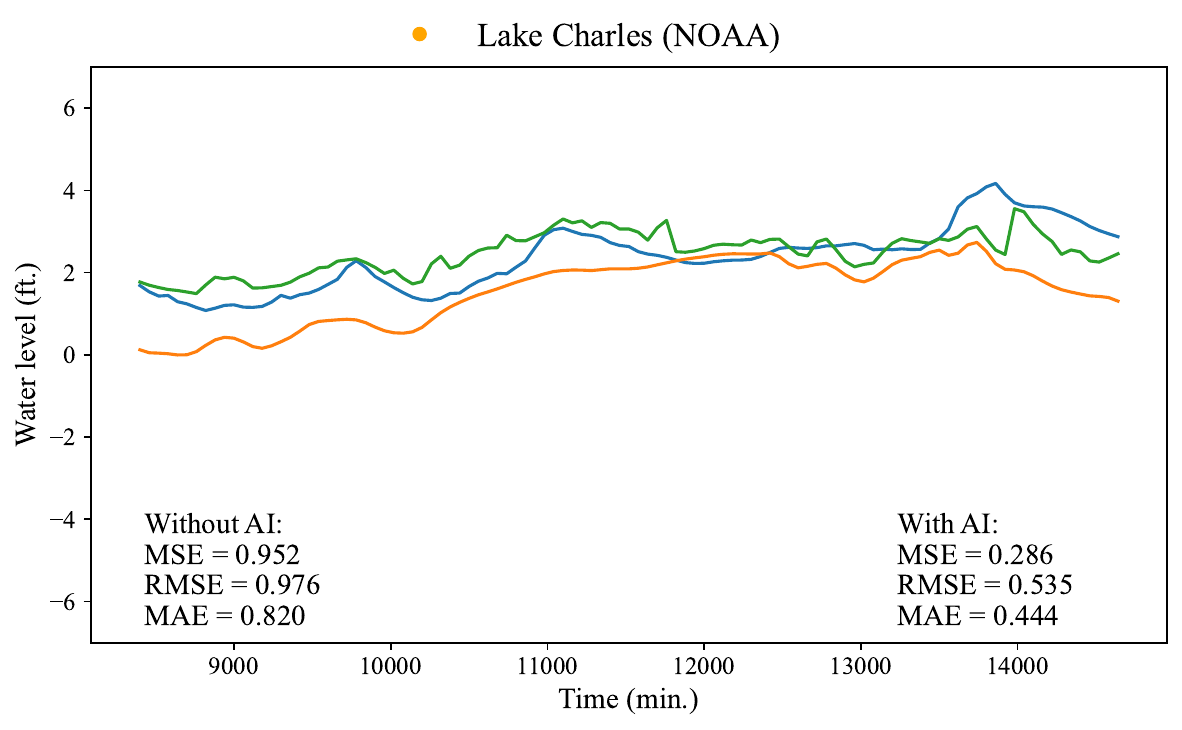}
        \end{subfigure}
        \begin{subfigure}[b]{0.25\textwidth}
            \centering
            \includegraphics[height=0.16\paperheight]{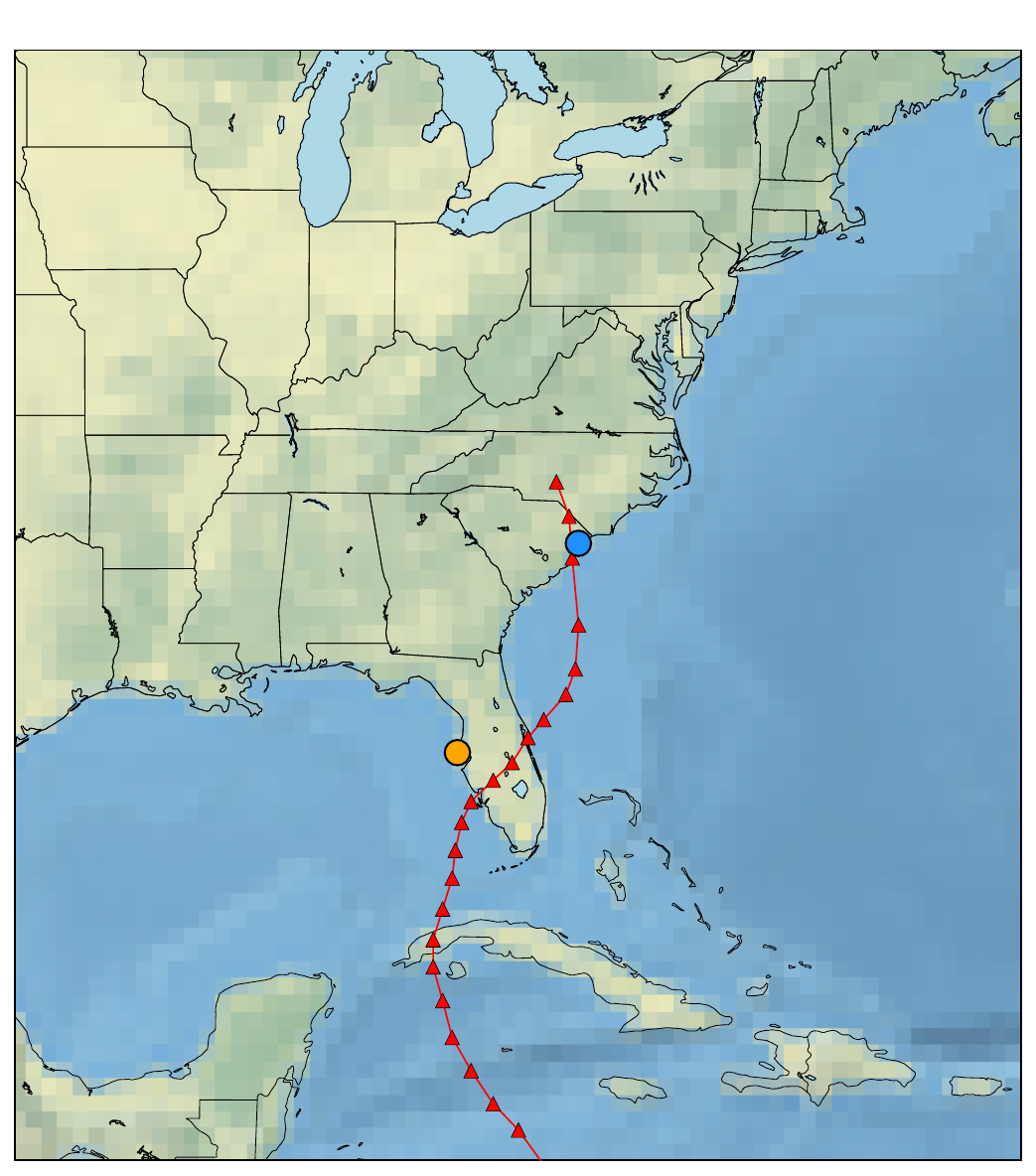}               
        \end{subfigure}
    \begin{subfigure}[b]{0.36\textwidth}
            \centering
            (c) Hurricane Ian (2022)
            \includegraphics[width=\textwidth]{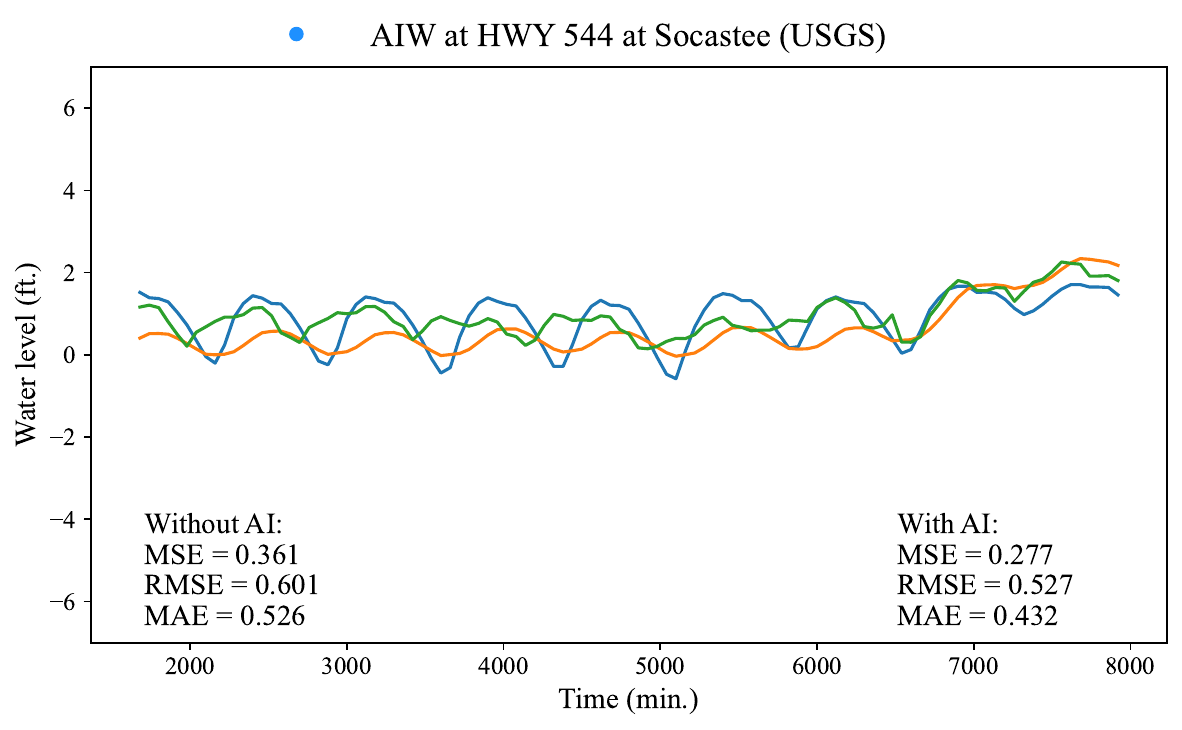}
        \end{subfigure}
    \begin{subfigure}[b]{0.36\textwidth}
            \centering
            \includegraphics[width=\textwidth]{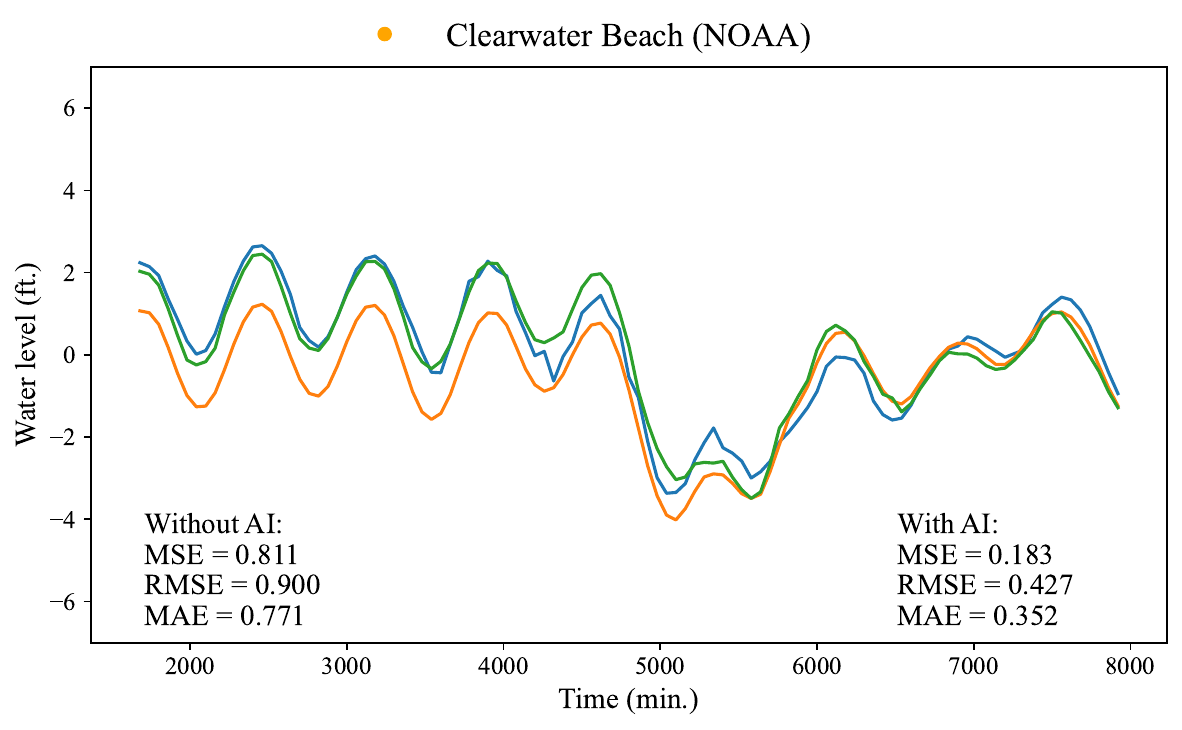}
        \end{subfigure}

    \caption{Comparison of observed (\textcolor{blue}{blue}), modeled (\textcolor{orange}{orange}) and \projectname-corrected modeled (\textcolor{ForestGreen}{green}) for: (a) Hermine (2016, category H1), (b) Harvey (2017, category H4), and (c) Ian (2022, category H5). The left side displays the location of two testing gauge stations (marked by colored disks) and the hurricane path (in red); while the right side shows the evaluation of the regression performance for these gauge stations. Evaluation statistics in each plot represent the performance of regression between modeled and observed water levels (without AI) and \projectname bias corrected and observed water levels (with AI).}
    \label{fig:station_wise}
\end{figure}

To evaluate how hurricane intensity impacts the extrapolating capabilities of \projectname, in Figure~\ref{fig:station_wise}, we demonstrate the effectiveness of the ML extrapolated water level corrections for two stations per each of three of the considered hurricanes, Hermine (2016), Harvey (2017) and Ian (2022), which lie towards the two ends of the Saffir-Simpson scale (H1, H4 and H5, respectively) \cite{SaffirSimpson}. Both stations for each hurricane are selected so to be in close geographical proximity to the corresponding storm track and simultaneously focusing on both heavily affected areas and less severely impacted regions.
By comparing the evaluation metrics of the non-bias corrected water level water level forecasts with the bias corrected ones with \projectname, it is evident that the \projectname extrapolation model is capable of improving predictions by reducing RMSE by $\sim 0.2-1.1$ ft. ($0.06-0.33$ m) (Figure~\ref{fig:station_wise}). This behavior is consistent regardless of storm intensity, although generated bias corrections are slightly less pronounced for the highest intensity hurricane (Hurricane Ian, 2022, H5). Moreover, the model can produce adequate corrections for both areas with high storm surge (e.g., USCG station Hatteras (NOAA) during Hurricane Hermine (2016) Figure~\ref{fig:station_wise}a) and less severely affected locations (e.g., Clearwater Beach (NOAA) during Hurricane Ian (2022), Figure~\ref{fig:station_wise}c).

\subsection{Inference Performance Analysis for Extrapolation with \projectname}

\begin{figure}
    \centering
    \includegraphics[width=0.7\linewidth]{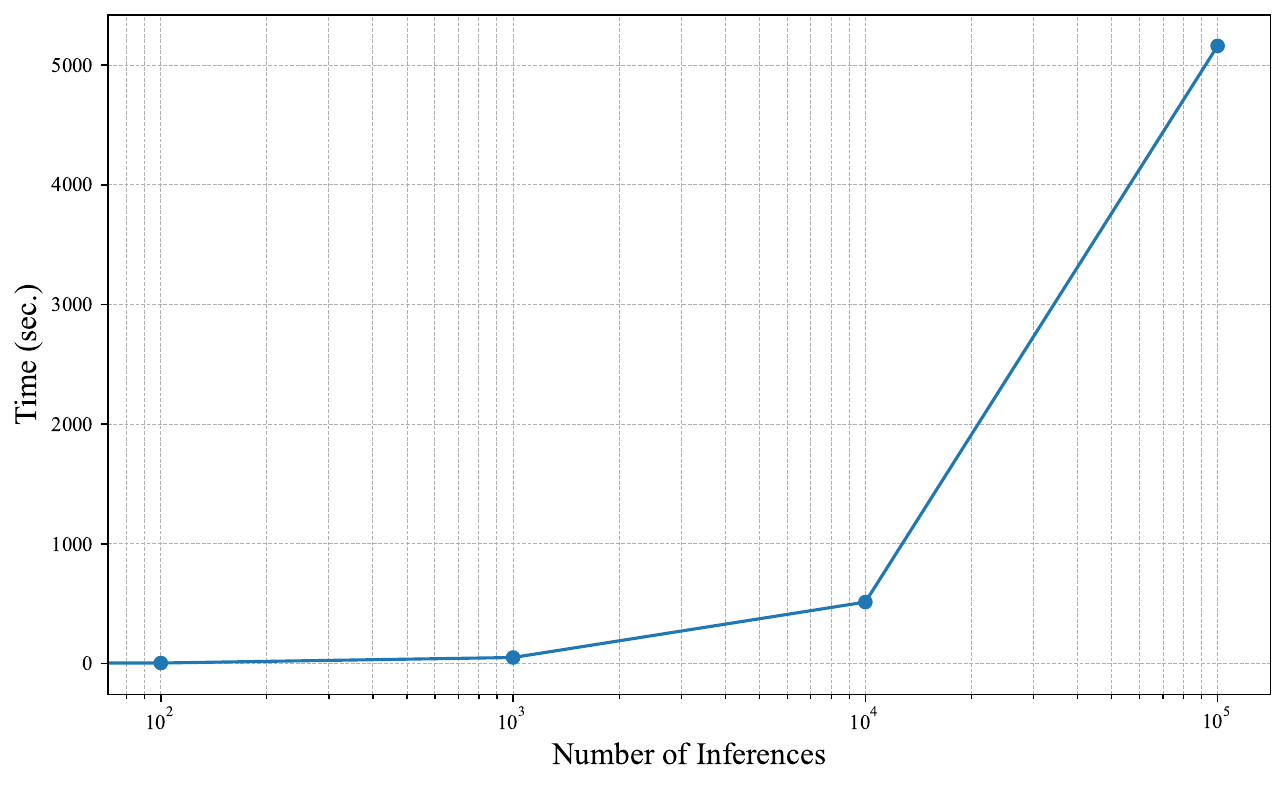}
    \caption{Computational time (in seconds) required for the \projectname model to sequentially extrapolate bias corrections (i.e., inference time), as a function of the total amount of corrections.}
    
    \label{fig:inference_analysis}
\end{figure}

In real-world applications, the ability to quickly extrapolate storm surge predictions for a large number of locations is critical. Therefore, an important aspect of evaluating \projectname is analyzing the time required for inference, which directly impacts the feasibility of using the model in 
the context of operational forecasting systems. In Figure \ref{fig:inference_analysis}, 
we present the computational time required for the \projectname model to generate inferences as a function of the amount of coordinates 
As expected, inference time increases with the number of coordinates at an exponential-like rate. Despite that a noticeable increase in inference time is observed when the model is required to generate more than $10^4$ values, the total inference time for $10^5$ values remains at a reasonable time of 5160 seconds (i.e., 1 h 40 min) and can potentially be further reduced by implementing additional parallelization schemes. This amount of data points could produce a high density coverage of a computational mesh spanning a wide enough region of interest during a storm. Moreover, given that forecasts from physics-based models (e.g., ADCIRC) are typically produced every 6 hours or so and station-wise offset prediction models such as our previous LSTM-based approach~\cite{Giaremis2024} have an almost negligible inference time, the inference time required for extrapolating offsets with \projectname is reasonable to allow its use as an additional bias correction component in an operational forecasting framework.

\section{Conclusion}
In this work, we explore the use of Generative AI (GenAI) methods through \projectname (Hurricane Bias Correction Beyond Gauge Stations Using Generative Adversarial Networks) to address the challenge of extrapolating bias-corrected water-level predictions to arbitrary geographic coordinates in storm surge modeling. \projectname is intended to be used in conjunction with a post-processing, station-wise bias estimator, such as our previously proposed LSTM-based model~\cite{Giaremis2024}, for estimating bias beyond gauge station locations. With \projectname, we introduce the use of a generative model based on the TimeGAN approach~\cite{yoon2019time} to extrapolate the behavior of the systemic error in storm tide forecast models, thereby enhancing the forecasting accuracy in areas without reference gauge stations during hurricane events. 

The dataset from the Historical Storm Surge Archive \cite{CERA2023} is used for the training and evaluating the proposed TimeGAN model. This dataset includes storm tide forecasts generated by ADCIRC \cite{Dietrich2011}, and observed water level data obtained from gauge stations. Here we consider data from Hurricanes Matthew (2016), Hermine (2016), Harvey (2017), Ida (2021), Ian (2022) and Idalia (2023).
The offset time series for all hurricanes are initially extracted using Eq.~\ref{eq:offsets}. Subsequently, we employ our previously developed LSTM-based model (\cite{Giaremis2024}) to generate offset prediction signals for each station and each of the considered hurricane. 
The stations are then clustered into groups, with one station per group designated for the test set and the remaining stations included in the training set. Subsequently, the data undergo normalization, cleaning, and reshaping to prepare it for input into \projectname. The TimeGAN-based structure and model parameters are finally chosen after hypertuning for each hurricane. We demonstrate that using an architecture with 256 neurons, 3000 epochs and 4 or 5 layers leads to the most accurate predictions in most cases, so it would be recommended for a real-world scenario. 
Our results show that our methodology can accurately extrapolate previously station-wise generated offsets at testing gauge station locations, unknown to the model during training, with satisfactory accuracy. Applying extrapolated bias corrections can consistently lead to a reduction of forecast RMSE, with little influence of the hurricane intensity. Inference times remain reasonable even for a large number of inferences (1h and 40 mins for $10^5$ inferences).



\textbf{Limitations and future work:} Limited performance is observed in few cases for stations along hurricane tracks or in locations further inland from the coastline or impacted by subtle coastline features. Therefore future work could focus on enhancing the robustness of the model by incorporating more diverse datasets, including data from different types of storms and geographical regions. Moreover, additional optimization could further improve inference times. Finally, implementing \projectname into real-time forecasting and evaluating its performance in operational settings would also be valuable directions for future research.

\section{Acknowledgments}
We acknowledge the support of the Department of Energy (DoE) through the award DE-SC0022320 (MuSiKAL). We would also like to thank Louisiana State University (LSU) and the Center for Computation and Technology at LSU for granting allocations for their computing resources and storage space.


\printcredits

\bibliographystyle{unsrtnat}
\bibliography{references}


\begin{thebibliography}{101}
\ifx \bisbn   \undefined \def \bisbn  #1{ISBN #1}\fi
\ifx \binits  \undefined \def \binits#1{#1}\fi
\ifx \bauthor  \undefined \def \bauthor#1{#1}\fi
\ifx \batitle  \undefined \def \batitle#1{#1}\fi
\ifx \bjtitle  \undefined \def \bjtitle#1{#1}\fi
\ifx \bvolume  \undefined \def \bvolume#1{\textbf{#1}}\fi
\ifx \byear  \undefined \def \byear#1{#1}\fi
\ifx \bissue  \undefined \def \bissue#1{#1}\fi
\ifx \bfpage  \undefined \def \bfpage#1{#1}\fi
\ifx \blpage  \undefined \def \blpage #1{#1}\fi
\ifx \burl  \undefined \def \burl#1{\textsf{#1}}\fi
\ifx \doiurl  \undefined \def \doiurl#1{\url{https://doi.org/#1}}\fi
\ifx \betal  \undefined \def \betal{\textit{et al.}}\fi
\ifx \binstitute  \undefined \def \binstitute#1{#1}\fi
\ifx \binstitutionaled  \undefined \def \binstitutionaled#1{#1}\fi
\ifx \bctitle  \undefined \def \bctitle#1{#1}\fi
\ifx \beditor  \undefined \def \beditor#1{#1}\fi
\ifx \bpublisher  \undefined \def \bpublisher#1{#1}\fi
\ifx \bbtitle  \undefined \def \bbtitle#1{#1}\fi
\ifx \bedition  \undefined \def \bedition#1{#1}\fi
\ifx \bseriesno  \undefined \def \bseriesno#1{#1}\fi
\ifx \blocation  \undefined \def \blocation#1{#1}\fi
\ifx \bsertitle  \undefined \def \bsertitle#1{#1}\fi
\ifx \bsnm \undefined \def \bsnm#1{#1}\fi
\ifx \bsuffix \undefined \def \bsuffix#1{#1}\fi
\ifx \bparticle \undefined \def \bparticle#1{#1}\fi
\ifx \barticle \undefined \def \barticle#1{#1}\fi
\bibcommenthead
\ifx \bconfdate \undefined \def \bconfdate #1{#1}\fi
\ifx \botherref \undefined \def \botherref #1{#1}\fi
\ifx \url \undefined \def \url#1{\textsf{#1}}\fi
\ifx \bchapter \undefined \def \bchapter#1{#1}\fi
\ifx \bbook \undefined \def \bbook#1{#1}\fi
\ifx \bcomment \undefined \def \bcomment#1{#1}\fi
\ifx \oauthor \undefined \def \oauthor#1{#1}\fi
\ifx \citeauthoryear \undefined \def \citeauthoryear#1{#1}\fi
\ifx \endbibitem  \undefined \def \endbibitem {}\fi
\ifx \bconflocation  \undefined \def \bconflocation#1{#1}\fi
\ifx \arxivurl  \undefined \def \arxivurl#1{\textsf{#1}}\fi
\csname PreBibitemsHook\endcsname

\bibitem[\protect\citeauthoryear{Smith}{2020}]{NCEI}
\begin{botherref}
\oauthor{\bsnm{Smith}, \binits{A.B.}}:
{U.S.} Billion-dollar Weather and Climate Disasters, 1980 - present ({NCEI Accession} 0209268). {NOAA National Centers for Environmental Information}. Dataset
(2020).
\doiurl{10.25921/STKW-7W73} .
\url{https://www.ncei.noaa.gov/access/billions/summary-stats/US/2004-2024}
\end{botherref}
\endbibitem

\bibitem[\protect\citeauthoryear{Calafat et~al.}{2022}]{Calafat2022}
\begin{barticle}
\bauthor{\bsnm{Calafat}, \binits{F.M.}},
\bauthor{\bsnm{Wahl}, \binits{T.}},
\bauthor{\bsnm{Tadesse}, \binits{M.G.}},
\bauthor{\bsnm{Sparrow}, \binits{S.N.}}:
\batitle{Trends in europe storm surge extremes match the rate of sea-level rise}.
\bjtitle{Nature}
\bvolume{603}(\bissue{7903}),
\bfpage{841}--\blpage{845}
(\byear{2022})
\doiurl{10.1038/s41586-022-04426-5}
\end{barticle}
\endbibitem

\bibitem[\protect\citeauthoryear{Wang and Toumi}{2021}]{Wang2021}
\begin{barticle}
\bauthor{\bsnm{Wang}, \binits{S.}},
\bauthor{\bsnm{Toumi}, \binits{R.}}:
\batitle{Recent migration of tropical cyclones toward coasts}.
\bjtitle{Science}
\bvolume{371}(\bissue{6528}),
\bfpage{514}--\blpage{517}
(\byear{2021})
\doiurl{10.1126/science.abb9038}
\end{barticle}
\endbibitem

\bibitem[\protect\citeauthoryear{Hall and Kossin}{2019}]{Hall2019}
\begin{botherref}
\oauthor{\bsnm{Hall}, \binits{T.M.}},
\oauthor{\bsnm{Kossin}, \binits{J.P.}}:
Hurricane stalling along the {North American} coast and implications for rainfall.
npj Climate and Atmospheric Science
\textbf{2}(1)
(2019)
\doiurl{10.1038/s41612-019-0074-8}
\end{botherref}
\endbibitem

\bibitem[\protect\citeauthoryear{Patricola and Wehner}{2018}]{Patricola2018}
\begin{barticle}
\bauthor{\bsnm{Patricola}, \binits{C.M.}},
\bauthor{\bsnm{Wehner}, \binits{M.F.}}:
\batitle{Anthropogenic influences on major tropical cyclone events}.
\bjtitle{Nature}
\bvolume{563}(\bissue{7731}),
\bfpage{339}--\blpage{346}
(\byear{2018})
\doiurl{10.1038/s41586-018-0673-2}
\end{barticle}
\endbibitem

\bibitem[\protect\citeauthoryear{Li and Chakraborty}{2020}]{Li2020}
\begin{barticle}
\bauthor{\bsnm{Li}, \binits{L.}},
\bauthor{\bsnm{Chakraborty}, \binits{P.}}:
\batitle{Slower decay of landfalling hurricanes in a warming world}.
\bjtitle{Nature}
\bvolume{587}(\bissue{7833}),
\bfpage{230}--\blpage{234}
(\byear{2020})
\doiurl{10.1038/s41586-020-2867-7}
\end{barticle}
\endbibitem

\bibitem[\protect\citeauthoryear{Balaguru et~al.}{2024}]{Balaguru2024}
\begin{botherref}
\oauthor{\bsnm{Balaguru}, \binits{K.}},
\oauthor{\bsnm{Chang}, \binits{C.}},
\oauthor{\bsnm{Leung}, \binits{L.R.}},
\oauthor{\bsnm{Foltz}, \binits{G.R.}},
\oauthor{\bsnm{Hagos}, \binits{S.M.}},
\oauthor{\bsnm{Wehner}, \binits{M.F.}},
\oauthor{\bsnm{Kossin}, \binits{J.P.}},
\oauthor{\bsnm{Ting}, \binits{M.}},
\oauthor{\bsnm{Xu}, \binits{W.}}:
A global increase in nearshore tropical cyclone intensification.
Earth's Future
\textbf{12}(5)
(2024)
\doiurl{10.1029/2023ef004230}
\end{botherref}
\endbibitem

\bibitem[\protect\citeauthoryear{Li et~al.}{2023}]{Li2023}
\begin{barticle}
\bauthor{\bsnm{Li}, \binits{X.}},
\bauthor{\bsnm{Zhan}, \binits{R.}},
\bauthor{\bsnm{Wang}, \binits{Y.}},
\bauthor{\bsnm{Zhao}, \binits{J.}},
\bauthor{\bsnm{Ding}, \binits{Y.}},
\bauthor{\bsnm{Song}, \binits{K.}}:
\batitle{Recent increase in rapid intensification events of tropical cyclones along {China} coast}.
\bjtitle{Climate Dynamics}
\bvolume{62}(\bissue{1}),
\bfpage{331}--\blpage{344}
(\byear{2023})
\doiurl{10.1007/s00382-023-06917-1}
\end{barticle}
\endbibitem

\bibitem[\protect\citeauthoryear{Shi et~al.}{2025}]{shi2025intensification}
\begin{botherref}
\oauthor{\bsnm{Shi}, \binits{J.}},
\oauthor{\bsnm{Hu}, \binits{C.}},
\oauthor{\bsnm{Cannizzaro}, \binits{J.}},
\oauthor{\bsnm{Barnes}, \binits{B.B.}},
\oauthor{\bsnm{Zhang}, \binits{Y.}},
\oauthor{\bsnm{Lembke}, \binits{C.}},
\oauthor{\bsnm{Le~Henaff}, \binits{M.}}:
Intensification of hurricane idalia by a river plume in the eastern gulf of mexico.
Environmental Research Letters
(2025)
\end{botherref}
\endbibitem

\bibitem[\protect\citeauthoryear{Liu et~al.}{2024}]{Liu2024}
\begin{botherref}
\oauthor{\bsnm{Liu}, \binits{Y.}},
\oauthor{\bsnm{Weisberg}, \binits{R.H.}},
\oauthor{\bsnm{Sorinas}, \binits{L.}},
\oauthor{\bsnm{Law}, \binits{J.A.}},
\oauthor{\bsnm{Nickerson}, \binits{A.K.}}:
Rapid intensification of {Hurricane Ian} in relation to anomalously warm subsurface water on the wide continental shelf.
Geophysical Research Letters
\textbf{52}(1)
(2024)
\doiurl{10.1029/2024gl113192}
\end{botherref}
\endbibitem

\bibitem[\protect\citeauthoryear{Zhu et~al.}{2022}]{Zhu2022}
\begin{barticle}
\bauthor{\bsnm{Zhu}, \binits{Y.-J.}},
\bauthor{\bsnm{Collins}, \binits{J.M.}},
\bauthor{\bsnm{Klotzbach}, \binits{P.J.}},
\bauthor{\bsnm{Schreck}, \binits{C.J.}}:
\batitle{{Hurricane Ida} (2021): Rapid intensification followed by slow inland decay}.
\bjtitle{Bulletin of the American Meteorological Society}
\bvolume{103}(\bissue{10}),
\bfpage{2354}--\blpage{2369}
(\byear{2022})
\doiurl{10.1175/bams-d-21-0240.1}
\end{barticle}
\endbibitem

\bibitem[\protect\citeauthoryear{Kotal et~al.}{2024}]{Kotal2024}
\begin{barticle}
\bauthor{\bsnm{Kotal}, \binits{S.D.}},
\bauthor{\bsnm{Arulalan}, \binits{T.}},
\bauthor{\bsnm{Mohapatra}, \binits{M.}}:
\batitle{Forecasting of tropical cyclones {ASANI} (2022) and {MOCHA} (2023) over the {Bay of Bengal} - real time challenges to forecasters}.
\bjtitle{Tropical Cyclone Research and Review}
\bvolume{13}(\bissue{2}),
\bfpage{88}--\blpage{112}
(\byear{2024})
\doiurl{10.1016/j.tcrr.2024.06.002}
\end{barticle}
\endbibitem

\bibitem[\protect\citeauthoryear{Petilla et~al.}{2025}]{Petilla2025}
\begin{barticle}
\bauthor{\bsnm{Petilla}, \binits{C.E.R.}},
\bauthor{\bsnm{Olaguera}, \binits{L.M.P.}},
\bauthor{\bsnm{Cruz}, \binits{F.A.T.}},
\bauthor{\bsnm{Villarin}, \binits{J.R.T.}},
\bauthor{\bsnm{Fudeyasu}, \binits{H.}},
\bauthor{\bsnm{Yoshida}, \binits{R.}},
\bauthor{\bsnm{Matsumoto}, \binits{J.}}:
\batitle{The unique features of typhoon rai (2021): an observational study}.
\bjtitle{Natural Hazards}
\bvolume{121}(\bissue{7}),
\bfpage{8279}--\blpage{8303}
(\byear{2025})
\doiurl{10.1007/s11069-025-07138-x}
\end{barticle}
\endbibitem

\bibitem[\protect\citeauthoryear{Zhang et~al.}{2019}]{Zhang2019}
\begin{barticle}
\bauthor{\bsnm{Zhang}, \binits{Z.}},
\bauthor{\bsnm{Wang}, \binits{Y.}},
\bauthor{\bsnm{Zhang}, \binits{W.}},
\bauthor{\bsnm{Xu}, \binits{J.}}:
\batitle{Coastal ocean response and its feedback to {Typhoon Hato} (2017) over the {South China Sea}: A numerical study}.
\bjtitle{Journal of Geophysical Research: Atmospheres}
\bvolume{124}(\bissue{24}),
\bfpage{13731}--\blpage{13749}
(\byear{2019})
\doiurl{10.1029/2019jd031377}
\end{barticle}
\endbibitem

\bibitem[\protect\citeauthoryear{Chang and Wu}{2017}]{Chang2017}
\begin{barticle}
\bauthor{\bsnm{Chang}, \binits{C.-C.}},
\bauthor{\bsnm{Wu}, \binits{C.-C.}}:
\batitle{On the processes leading to the rapid intensification of {Typhoon Megi} (2010)}.
\bjtitle{Journal of the Atmospheric Sciences}
\bvolume{74}(\bissue{4}),
\bfpage{1169}--\blpage{1200}
(\byear{2017})
\doiurl{10.1175/jas-d-16-0075.1}
\end{barticle}
\endbibitem

\bibitem[\protect\citeauthoryear{Wu et~al.}{2015}]{Wu2015}
\begin{barticle}
\bauthor{\bsnm{Wu}, \binits{L.}},
\bauthor{\bsnm{Su}, \binits{H.}},
\bauthor{\bsnm{Fovell}, \binits{R.G.}},
\bauthor{\bsnm{Dunkerton}, \binits{T.J.}},
\bauthor{\bsnm{Wang}, \binits{Z.}},
\bauthor{\bsnm{Kahn}, \binits{B.H.}}:
\batitle{Impact of environmental moisture on tropical cyclone intensification}.
\bjtitle{Atmospheric Chemistry and Physics}
\bvolume{15}(\bissue{24}),
\bfpage{14041}--\blpage{14053}
(\byear{2015})
\doiurl{10.5194/acp-15-14041-2015}
\end{barticle}
\endbibitem

\bibitem[\protect\citeauthoryear{Zagrodnik and Jiang}{2014}]{Zagrodnik2014}
\begin{barticle}
\bauthor{\bsnm{Zagrodnik}, \binits{J.P.}},
\bauthor{\bsnm{Jiang}, \binits{H.}}:
\batitle{Rainfall, convection, and latent heating distributions in rapidly intensifying tropical cyclones}.
\bjtitle{Journal of the Atmospheric Sciences}
\bvolume{71}(\bissue{8}),
\bfpage{2789}--\blpage{2809}
(\byear{2014})
\doiurl{10.1175/jas-d-13-0314.1}
\end{barticle}
\endbibitem

\bibitem[\protect\citeauthoryear{Wu et~al.}{2025}]{Wu2025}
\begin{botherref}
\oauthor{\bsnm{Wu}, \binits{X.}},
\oauthor{\bsnm{Hoffmann}, \binits{L.}},
\oauthor{\bsnm{Wright}, \binits{C.J.}},
\oauthor{\bsnm{Hindley}, \binits{N.P.}},
\oauthor{\bsnm{Alexander}, \binits{M.J.}},
\oauthor{\bsnm{Wang}, \binits{X.}},
\oauthor{\bsnm{Chen}, \binits{B.}},
\oauthor{\bsnm{Wang}, \binits{Y.}},
\oauthor{\bsnm{Li}, \binits{M.}}:
Mechanisms linking stratospheric gravity wave activity to hurricane intensification: Insights from model simulation of {Hurricane Joaquin}.
Geophysical Research Letters
\textbf{52}(10)
(2025)
\doiurl{10.1029/2024gl113531}
\end{botherref}
\endbibitem

\bibitem[\protect\citeauthoryear{Yang et~al.}{2024}]{Yang2024}
\begin{barticle}
\bauthor{\bsnm{Yang}, \binits{S.}},
\bauthor{\bsnm{Shin}, \binits{D.}},
\bauthor{\bsnm{Cocke}, \binits{S.}},
\bauthor{\bsnm{Nam}, \binits{C.C.}},
\bauthor{\bsnm{Bourassa}, \binits{M.}},
\bauthor{\bsnm{Cha}, \binits{D.-H.}},
\bauthor{\bsnm{Kim}, \binits{B.-M.}}:
\batitle{Unveiling the pivotal influence of sea spray heat fluxes on hurricane rapid intensification}.
\bjtitle{Environmental Research Letters}
\bvolume{19}(\bissue{11}),
\bfpage{114058}
(\byear{2024})
\doiurl{10.1088/1748-9326/ad7ee0}
\end{barticle}
\endbibitem

\bibitem[\protect\citeauthoryear{Kim et~al.}{2024}]{Kim2024}
\begin{botherref}
\oauthor{\bsnm{Kim}, \binits{J.-H.}},
\oauthor{\bsnm{Ham}, \binits{Y.-G.}},
\oauthor{\bsnm{Kim}, \binits{D.}},
\oauthor{\bsnm{Li}, \binits{T.}},
\oauthor{\bsnm{Ma}, \binits{C.}}:
Improvement in forecasting short-term tropical cyclone intensity change and their rapid intensification using deep learning.
Artificial Intelligence for the Earth Systems
\textbf{3}(2)
(2024)
\doiurl{10.1175/aies-d-23-0052.1}
\end{botherref}
\endbibitem

\bibitem[\protect\citeauthoryear{Cangialosi et~al.}{2020}]{Cangialosi2020}
\begin{barticle}
\bauthor{\bsnm{Cangialosi}, \binits{J.P.}},
\bauthor{\bsnm{Blake}, \binits{E.}},
\bauthor{\bsnm{DeMaria}, \binits{M.}},
\bauthor{\bsnm{Penny}, \binits{A.}},
\bauthor{\bsnm{Latto}, \binits{A.}},
\bauthor{\bsnm{Rappaport}, \binits{E.}},
\bauthor{\bsnm{Tallapragada}, \binits{V.}}:
\batitle{Recent progress in tropical cyclone intensity forecasting at the {National Hurricane Center}}.
\bjtitle{Weather and Forecasting}
\bvolume{35}(\bissue{5}),
\bfpage{1913}--\blpage{1922}
(\byear{2020})
\doiurl{10.1175/waf-d-20-0059.1}
\end{barticle}
\endbibitem

\bibitem[\protect\citeauthoryear{Trabing and Bell}{2020}]{Trabing2020}
\begin{barticle}
\bauthor{\bsnm{Trabing}, \binits{B.C.}},
\bauthor{\bsnm{Bell}, \binits{M.M.}}:
\batitle{Understanding error distributions of hurricane intensity forecasts during rapid intensity changes}.
\bjtitle{Weather and Forecasting}
\bvolume{35}(\bissue{6}),
\bfpage{2219}--\blpage{2234}
(\byear{2020})
\doiurl{10.1175/waf-d-19-0253.1}
\end{barticle}
\endbibitem

\bibitem[\protect\citeauthoryear{Cyriac et~al.}{2018}]{Cyriac2018}
\begin{barticle}
\bauthor{\bsnm{Cyriac}, \binits{R.}},
\bauthor{\bsnm{Dietrich}, \binits{J.C.}},
\bauthor{\bsnm{Fleming}, \binits{J.G.}},
\bauthor{\bsnm{Blanton}, \binits{B.O.}},
\bauthor{\bsnm{Kaiser}, \binits{C.}},
\bauthor{\bsnm{Dawson}, \binits{C.N.}},
\bauthor{\bsnm{Luettich}, \binits{R.A.}}:
\batitle{Variability in coastal flooding predictions due to forecast errors during {Hurricane} {Arthur}}.
\bjtitle{Coastal Engineering}
\bvolume{137},
\bfpage{59}--\blpage{78}
(\byear{2018})
\doiurl{10.1016/j.coastaleng.2018.02.008}
\end{barticle}
\endbibitem

\bibitem[\protect\citeauthoryear{Turner et~al.}{2024}]{Turner2024}
\begin{barticle}
\bauthor{\bsnm{Turner}, \binits{I.L.}},
\bauthor{\bsnm{Leaman}, \binits{C.K.}},
\bauthor{\bsnm{Harley}, \binits{M.D.}},
\bauthor{\bsnm{Thran}, \binits{M.C.}},
\bauthor{\bsnm{David}, \binits{D.R.}},
\bauthor{\bsnm{Splinter}, \binits{K.D.}},
\bauthor{\bsnm{Matheen}, \binits{N.}},
\bauthor{\bsnm{Hansen}, \binits{J.E.}},
\bauthor{\bsnm{Cuttler}, \binits{M.V.W.}},
\bauthor{\bsnm{Greenslade}, \binits{D.J.M.}},
\bauthor{\bsnm{Zieger}, \binits{S.}},
\bauthor{\bsnm{Lowe}, \binits{R.J.}}:
\batitle{A framework for national-scale coastal storm hazards early warning}.
\bjtitle{Coastal Engineering}
\bvolume{192},
\bfpage{104571}
(\byear{2024})
\doiurl{10.1016/j.coastaleng.2024.104571}
\end{barticle}
\endbibitem

\bibitem[\protect\citeauthoryear{Penny et~al.}{2023}]{Penny2023}
\begin{barticle}
\bauthor{\bsnm{Penny}, \binits{A.B.}},
\bauthor{\bsnm{Alaka}, \binits{L.}},
\bauthor{\bsnm{Taylor}, \binits{A.A.}},
\bauthor{\bsnm{Booth}, \binits{W.}},
\bauthor{\bsnm{DeMaria}, \binits{M.}},
\bauthor{\bsnm{Fritz}, \binits{C.}},
\bauthor{\bsnm{Rhome}, \binits{J.}}:
\batitle{Operational storm surge forecasting at the national hurricane center: The case for probabilistic guidance and the evaluation of improved storm size forecasts used to define the wind forcing}.
\bjtitle{Weather and Forecasting}
\bvolume{38}(\bissue{12}),
\bfpage{2461}--\blpage{2479}
(\byear{2023})
\doiurl{10.1175/waf-d-22-0209.1}
\end{barticle}
\endbibitem

\bibitem[\protect\citeauthoryear{Suh et~al.}{2015}]{Suh2015}
\begin{barticle}
\bauthor{\bsnm{Suh}, \binits{S.W.}},
\bauthor{\bsnm{Lee}, \binits{H.Y.}},
\bauthor{\bsnm{Kim}, \binits{H.J.}},
\bauthor{\bsnm{Fleming}, \binits{J.G.}}:
\batitle{An efficient early warning system for typhoon storm surge based on time-varying advisories by coupled {ADCIRC} and {SWAN}}.
\bjtitle{Ocean Dynamics}
\bvolume{65}(\bissue{5}),
\bfpage{617}--\blpage{646}
(\byear{2015})
\doiurl{10.1007/s10236-015-0820-3}
\end{barticle}
\endbibitem

\bibitem[\protect\citeauthoryear{Westerink et~al.}{1992}]{Westerink1992}
\begin{barticle}
\bauthor{\bsnm{Westerink}, \binits{J.J.}},
\bauthor{\bsnm{Luettich}, \binits{R.A.}},
\bauthor{\bsnm{Baptists}, \binits{A.M.}},
\bauthor{\bsnm{Scheffner}, \binits{N.W.}},
\bauthor{\bsnm{Farrar}, \binits{P.}}:
\batitle{Tide and storm surge predictions using finite element model}.
\bjtitle{Journal of Hydraulic Engineering}
\bvolume{118}(\bissue{10}),
\bfpage{1373}--\blpage{1390}
(\byear{1992})
\doiurl{10.1061/(asce)0733-9429(1992)118:10(1373)}
\end{barticle}
\endbibitem

\bibitem[\protect\citeauthoryear{Booij et~al.}{1999}]{Booij1999}
\begin{barticle}
\bauthor{\bsnm{Booij}, \binits{N.}},
\bauthor{\bsnm{Ris}, \binits{R.C.}},
\bauthor{\bsnm{Holthuijsen}, \binits{L.H.}}:
\batitle{A third‐generation wave model for coastal regions: 1. model description and validation}.
\bjtitle{Journal of Geophysical Research: Oceans}
\bvolume{104}(\bissue{C4}),
\bfpage{7649}--\blpage{7666}
(\byear{1999})
\doiurl{10.1029/98jc02622}
\end{barticle}
\endbibitem

\bibitem[\protect\citeauthoryear{Dietrich et~al.}{2011}]{Dietrich2011}
\begin{barticle}
\bauthor{\bsnm{Dietrich}, \binits{J.C.}},
\bauthor{\bsnm{Zijlema}, \binits{M.}},
\bauthor{\bsnm{Westerink}, \binits{J.J.}},
\bauthor{\bsnm{Holthuijsen}, \binits{L.H.}},
\bauthor{\bsnm{Dawson}, \binits{C.}},
\bauthor{\bsnm{Luettich}, \binits{R.A.}},
\bauthor{\bsnm{Jensen}, \binits{R.E.}},
\bauthor{\bsnm{Smith}, \binits{J.M.}},
\bauthor{\bsnm{Stelling}, \binits{G.S.}},
\bauthor{\bsnm{Stone}, \binits{G.W.}}:
\batitle{Modeling hurricane waves and storm surge using integrally-coupled, scalable computations}.
\bjtitle{Coastal Engineering}
\bvolume{58}(\bissue{1}),
\bfpage{45}--\blpage{65}
(\byear{2011})
\doiurl{10.1016/j.coastaleng.2010.08.001}
\end{barticle}
\endbibitem

\bibitem[\protect\citeauthoryear{{CERA - Coastal Emergency Risk Accessment}}{2025}]{CERA}
\begin{botherref}
\oauthor{\bsnm{{CERA - Coastal Emergency Risk Accessment}}}:
{Center for Computation and Technology at Louisiana State University}.
\url{https://cera.coastalrisk.live/}
(2025)
\end{botherref}
\endbibitem

\bibitem[\protect\citeauthoryear{Chen et~al.}{2025}]{Chen2025}
\begin{barticle}
\bauthor{\bsnm{Chen}, \binits{F.}},
\bauthor{\bsnm{Yang}, \binits{W.}},
\bauthor{\bsnm{Xiao}, \binits{L.}},
\bauthor{\bsnm{Xia}, \binits{X.}},
\bauthor{\bsnm{Ding}, \binits{K.}},
\bauthor{\bsnm{Sun}, \binits{Z.}}:
\batitle{An exploratory assessment of a submarine topographic characteristic index for predicting extreme flow velocities: A case study of {Typhoon In---Fa} in the {Zhoushan Sea} area}.
\bjtitle{Journal of Marine Science and Engineering}
\bvolume{13}(\bissue{5}),
\bfpage{864}
(\byear{2025})
\doiurl{10.3390/jmse13050864}
\end{barticle}
\endbibitem

\bibitem[\protect\citeauthoryear{Pringle et~al.}{2021}]{Pringle2021}
\begin{barticle}
\bauthor{\bsnm{Pringle}, \binits{W.J.}},
\bauthor{\bsnm{Wirasaet}, \binits{D.}},
\bauthor{\bsnm{Roberts}, \binits{K.J.}},
\bauthor{\bsnm{Westerink}, \binits{J.J.}}:
\batitle{Global storm tide modeling with {ADCIRC} v55: unstructured mesh design and performance}.
\bjtitle{Geoscientific Model Development}
\bvolume{14}(\bissue{2}),
\bfpage{1125}--\blpage{1145}
(\byear{2021})
\doiurl{10.5194/gmd-14-1125-2021}
\end{barticle}
\endbibitem

\bibitem[\protect\citeauthoryear{Khani and Dawson}{2023}]{Khani2023}
\begin{botherref}
\oauthor{\bsnm{Khani}, \binits{S.}},
\oauthor{\bsnm{Dawson}, \binits{C.N.}}:
A gradient based subgrid‐scale parameterization for ocean mesoscale eddies.
Journal of Advances in Modeling Earth Systems
\textbf{15}(2)
(2023)
\doiurl{10.1029/2022ms003356}
\end{botherref}
\endbibitem

\bibitem[\protect\citeauthoryear{Blakely et~al.}{2022}]{Blakely2022}
\begin{botherref}
\oauthor{\bsnm{Blakely}, \binits{C.P.}},
\oauthor{\bsnm{Ling}, \binits{G.}},
\oauthor{\bsnm{Pringle}, \binits{W.J.}},
\oauthor{\bsnm{Contreras}, \binits{M.T.}},
\oauthor{\bsnm{Wirasaet}, \binits{D.}},
\oauthor{\bsnm{Westerink}, \binits{J.J.}},
\oauthor{\bsnm{Moghimi}, \binits{S.}},
\oauthor{\bsnm{Seroka}, \binits{G.}},
\oauthor{\bsnm{Shi}, \binits{L.}},
\oauthor{\bsnm{Myers}, \binits{E.}},
\oauthor{\bsnm{Owensby}, \binits{M.}},
\oauthor{\bsnm{Massey}, \binits{C.}}:
Dissipation and bathymetric sensitivities in an unstructured mesh global tidal model.
Journal of Geophysical Research: Oceans
\textbf{127}(5)
(2022)
\doiurl{10.1029/2021jc018178}
\end{botherref}
\endbibitem

\bibitem[\protect\citeauthoryear{Loveland et~al.}{2024}]{Loveland2024}
\begin{barticle}
\bauthor{\bsnm{Loveland}, \binits{M.}},
\bauthor{\bsnm{Meixner}, \binits{J.}},
\bauthor{\bsnm{Valseth}, \binits{E.}},
\bauthor{\bsnm{Dawson}, \binits{C.}}:
\batitle{Efficacy of reduced order source terms for a coupled wave-circulation model in the {Gulf} of {Mexico}}.
\bjtitle{Ocean Modelling}
\bvolume{190},
\bfpage{102387}
(\byear{2024})
\doiurl{10.1016/j.ocemod.2024.102387}
\end{barticle}
\endbibitem

\bibitem[\protect\citeauthoryear{Dawson et~al.}{2024}]{Dawson2024}
\begin{botherref}
\oauthor{\bsnm{Dawson}, \binits{C.}},
\oauthor{\bsnm{Loveland}, \binits{M.}},
\oauthor{\bsnm{Pachev}, \binits{B.}},
\oauthor{\bsnm{Proft}, \binits{J.}},
\oauthor{\bsnm{Valseth}, \binits{E.}}:
{SWEMniCS}: a software toolbox for modeling coastal ocean circulation, storm surges, inland, and compound flooding.
npj Natural Hazards
\textbf{1}(1)
(2024)
\doiurl{10.1038/s44304-024-00036-5}
\end{botherref}
\endbibitem

\bibitem[\protect\citeauthoryear{Bernier et~al.}{2024}]{Bernier2024}
\begin{barticle}
\bauthor{\bsnm{Bernier}, \binits{N.B.}},
\bauthor{\bsnm{Hemer}, \binits{M.}},
\bauthor{\bsnm{Mori}, \binits{N.}},
\bauthor{\bsnm{Appendini}, \binits{C.M.}},
\bauthor{\bsnm{Breivik}, \binits{O.}},
\bauthor{\bsnm{Camargo}, \binits{R.}},
\bauthor{\bsnm{Casas-Prat}, \binits{M.}},
\bauthor{\bsnm{Duong}, \binits{T.M.}},
\bauthor{\bsnm{Haigh}, \binits{I.D.}},
\bauthor{\bsnm{Howard}, \binits{T.}},
\bauthor{\bsnm{Hernaman}, \binits{V.}},
\bauthor{\bsnm{Huizy}, \binits{O.}},
\bauthor{\bsnm{Irish}, \binits{J.L.}},
\bauthor{\bsnm{Kirezci}, \binits{E.}},
\bauthor{\bsnm{Kohno}, \binits{N.}},
\bauthor{\bsnm{Lee}, \binits{J.-W.}},
\bauthor{\bsnm{McInnes}, \binits{K.L.}},
\bauthor{\bsnm{Meyer}, \binits{E.M.I.}},
\bauthor{\bsnm{Marcos}, \binits{M.}},
\bauthor{\bsnm{Marsooli}, \binits{R.}},
\bauthor{\bsnm{Martin~Oliva}, \binits{A.}},
\bauthor{\bsnm{Menendez}, \binits{M.}},
\bauthor{\bsnm{Moghimi}, \binits{S.}},
\bauthor{\bsnm{Muis}, \binits{S.}},
\bauthor{\bsnm{Polton}, \binits{J.A.}},
\bauthor{\bsnm{Pringle}, \binits{W.J.}},
\bauthor{\bsnm{Ranasinghe}, \binits{R.}},
\bauthor{\bsnm{Saillour}, \binits{T.}},
\bauthor{\bsnm{Smith}, \binits{G.}},
\bauthor{\bsnm{Tadesse}, \binits{M.G.}},
\bauthor{\bsnm{Swail}, \binits{V.}},
\bauthor{\bsnm{Tomoya}, \binits{S.}},
\bauthor{\bsnm{Voukouvalas}, \binits{E.}},
\bauthor{\bsnm{Wahl}, \binits{T.}},
\bauthor{\bsnm{Wang}, \binits{P.}},
\bauthor{\bsnm{Weisse}, \binits{R.}},
\bauthor{\bsnm{Westerink}, \binits{J.J.}},
\bauthor{\bsnm{Young}, \binits{I.}},
\bauthor{\bsnm{Zhang}, \binits{Y.J.}}:
\batitle{Storm surges and extreme sea levels: Review, establishment of model intercomparison and coordination of surge climate projection efforts ({SurgeMIP}).}
\bjtitle{Weather and Climate Extremes}
\bvolume{45},
\bfpage{100689}
(\byear{2024})
\doiurl{10.1016/j.wace.2024.100689}
\end{barticle}
\endbibitem

\bibitem[\protect\citeauthoryear{Loveland et~al.}{2021}]{Loveland2021}
\begin{botherref}
\oauthor{\bsnm{Loveland}, \binits{M.}},
\oauthor{\bsnm{Kiaghadi}, \binits{A.}},
\oauthor{\bsnm{Dawson}, \binits{C.N.}},
\oauthor{\bsnm{Rifai}, \binits{H.S.}},
\oauthor{\bsnm{Misra}, \binits{S.}},
\oauthor{\bsnm{Mosser}, \binits{H.}},
\oauthor{\bsnm{Parola}, \binits{A.}}:
Developing a modeling framework to simulate compound flooding: When storm surge interacts with riverine flow.
Frontiers in Climate
\textbf{2}
(2021)
\doiurl{10.3389/fclim.2020.609610}
\end{botherref}
\endbibitem

\bibitem[\protect\citeauthoryear{Wei et~al.}{2024}]{Wei2024}
\begin{botherref}
\oauthor{\bsnm{Wei}, \binits{W.}},
\oauthor{\bsnm{Huang}, \binits{S.}},
\oauthor{\bsnm{Qin}, \binits{H.}},
\oauthor{\bsnm{Yu}, \binits{L.}},
\oauthor{\bsnm{Mu}, \binits{L.}}:
Storm surge risk assessment and sensitivity analysis based on multiple criteria decision-making methods: a case study of {Huizhou} city.
Frontiers in Marine Science
\textbf{11}
(2024)
\doiurl{10.3389/fmars.2024.1364929}
\end{botherref}
\endbibitem

\bibitem[\protect\citeauthoryear{Zhang et~al.}{2023}]{Zhang2023}
\begin{barticle}
\bauthor{\bsnm{Zhang}, \binits{Z.}},
\bauthor{\bsnm{Lu}, \binits{Y.}},
\bauthor{\bsnm{Hu}, \binits{D.}},
\bauthor{\bsnm{Guo}, \binits{F.}},
\bauthor{\bsnm{Yu}, \binits{Z.}},
\bauthor{\bsnm{Song}, \binits{Z.}},
\bauthor{\bsnm{Chen}, \binits{P.}},
\bauthor{\bsnm{Wu}, \binits{J.}},
\bauthor{\bsnm{Huang}, \binits{W.}}:
\batitle{A cross-scale modeling framework for simulating typhoon-induced compound floods and assessing the emergency response in urban regions}.
\bjtitle{Ocean \& Coastal Management}
\bvolume{245},
\bfpage{106863}
(\byear{2023})
\doiurl{10.1016/j.ocecoaman.2023.106863}
\end{barticle}
\endbibitem

\bibitem[\protect\citeauthoryear{Huang et~al.}{2021}]{Huang2021}
\begin{barticle}
\bauthor{\bsnm{Huang}, \binits{W.}},
\bauthor{\bsnm{Yin}, \binits{K.}},
\bauthor{\bsnm{Ghorbanzadeh}, \binits{M.}},
\bauthor{\bsnm{Ozguven}, \binits{E.}},
\bauthor{\bsnm{Xu}, \binits{S.}},
\bauthor{\bsnm{Vijayan}, \binits{L.}}:
\batitle{Integrating storm surge modeling with traffic data analysis to evaluate the effectiveness of hurricane evacuation}.
\bjtitle{Frontiers of Structural and Civil Engineering}
\bvolume{15}(\bissue{6}),
\bfpage{1301}--\blpage{1316}
(\byear{2021})
\doiurl{10.1007/s11709-021-0765-1}
\end{barticle}
\endbibitem

\bibitem[\protect\citeauthoryear{\"{O}zkan et~al.}{2025}]{Ozkan2025}
\begin{botherref}
\oauthor{\bsnm{\"{O}zkan}, \binits{F.N.}},
\oauthor{\bsnm{Verlaan}, \binits{M.}},
\oauthor{\bsnm{Muis}, \binits{S.}},
\oauthor{\bsnm{Zijl}, \binits{F.}}:
Sensitivity of global storm surge modelling to sea surface drag.
Ocean Dynamics
\textbf{75}(8)
(2025)
\doiurl{10.1007/s10236-025-01713-3}
\end{botherref}
\endbibitem

\bibitem[\protect\citeauthoryear{Muñoz et~al.}{2022}]{Munoz2022}
\begin{barticle}
\bauthor{\bsnm{Muñoz}, \binits{D.F.}},
\bauthor{\bsnm{Abbaszadeh}, \binits{P.}},
\bauthor{\bsnm{Moftakhari}, \binits{H.}},
\bauthor{\bsnm{Moradkhani}, \binits{H.}}:
\batitle{Accounting for uncertainties in compound flood hazard assessment: The value of data assimilation}.
\bjtitle{Coastal Engineering}
\bvolume{171},
\bfpage{104057}
(\byear{2022})
\doiurl{10.1016/j.coastaleng.2021.104057}
\end{barticle}
\endbibitem

\bibitem[\protect\citeauthoryear{Torres et~al.}{2019}]{Torres2019}
\begin{botherref}
\oauthor{\bsnm{Torres}, \binits{M.J.}},
\oauthor{\bsnm{Reza~Hashemi}, \binits{M.}},
\oauthor{\bsnm{Hayward}, \binits{S.}},
\oauthor{\bsnm{Spaulding}, \binits{M.}},
\oauthor{\bsnm{Ginis}, \binits{I.}},
\oauthor{\bsnm{Grilli}, \binits{S.T.}}:
Role of hurricane wind models in accurate simulation of storm surge and waves.
Journal of Waterway, Port, Coastal, and Ocean Engineering
\textbf{145}(1)
(2019)
\doiurl{10.1061/(asce)ww.1943-5460.0000496}
\end{botherref}
\endbibitem

\bibitem[\protect\citeauthoryear{Gallien et~al.}{2018}]{Gallien2018}
\begin{barticle}
\bauthor{\bsnm{Gallien}, \binits{T.W.}},
\bauthor{\bsnm{Kalligeris}, \binits{N.}},
\bauthor{\bsnm{Delisle}, \binits{M.-P.C.}},
\bauthor{\bsnm{Tang}, \binits{B.-X.}},
\bauthor{\bsnm{Lucey}, \binits{J.T.D.}},
\bauthor{\bsnm{Winters}, \binits{M.A.}}:
\batitle{Coastal flood modeling challenges in defended urban backshores}.
\bjtitle{Geosciences}
\bvolume{8}(\bissue{12}),
\bfpage{450}
(\byear{2018})
\doiurl{10.3390/geosciences8120450}
\end{barticle}
\endbibitem

\bibitem[\protect\citeauthoryear{Ferreira et~al.}{2014}]{Ferreira2014}
\begin{barticle}
\bauthor{\bsnm{Ferreira}, \binits{C.M.}},
\bauthor{\bsnm{Irish}, \binits{J.L.}},
\bauthor{\bsnm{Olivera}, \binits{F.}}:
\batitle{Uncertainty in hurricane surge simulation due to land cover specification}.
\bjtitle{Journal of Geophysical Research: Oceans}
\bvolume{119}(\bissue{3}),
\bfpage{1812}--\blpage{1827}
(\byear{2014})
\doiurl{10.1002/2013jc009604}
\end{barticle}
\endbibitem

\bibitem[\protect\citeauthoryear{Asher et~al.}{2019}]{Asher2019}
\begin{barticle}
\bauthor{\bsnm{Asher}, \binits{T.G.}},
\bauthor{\bsnm{Luettich~Jr.}, \binits{R.A.}},
\bauthor{\bsnm{Fleming}, \binits{J.G.}},
\bauthor{\bsnm{Blanton}, \binits{B.O.}}:
\batitle{Low frequency water level correction in storm surge models using data assimilation}.
\bjtitle{Ocean Modelling}
\bvolume{144},
\bfpage{101483}
(\byear{2019})
\doiurl{10.1016/j.ocemod.2019.101483}
\end{barticle}
\endbibitem

\bibitem[\protect\citeauthoryear{Gonzalez et~al.}{2019}]{Gonzalez2019}
\begin{botherref}
\oauthor{\bsnm{Gonzalez}, \binits{V.M.}},
\oauthor{\bsnm{Nadal-Caraballo}, \binits{N.C.}},
\oauthor{\bsnm{Melby}, \binits{J.A.}},
\oauthor{\bsnm{Cialone}, \binits{M.A.}}:
Quantification of uncertainty in probabilistic storm surge models: Literature review.
Technical report,
U.S. Army Corps of Engineers, Engineer Research and Development Center
(2019).
\url{https://chs.erdc.dren.mil/Library/References/CHS_PCHA_Publications/Reports/SR-19-1_Gonzalez_et_al_2019_UncertaintyInSurgeModels.pdf}
\end{botherref}
\endbibitem

\bibitem[\protect\citeauthoryear{Resio et~al.}{2012}]{Resio2012}
\begin{barticle}
\bauthor{\bsnm{Resio}, \binits{D.T.}},
\bauthor{\bsnm{Irish}, \binits{J.L.}},
\bauthor{\bsnm{Westerink}, \binits{J.J.}},
\bauthor{\bsnm{Powell}, \binits{N.J.}}:
\batitle{The effect of uncertainty on estimates of hurricane surge hazards}.
\bjtitle{Natural Hazards}
\bvolume{66}(\bissue{3}),
\bfpage{1443}--\blpage{1459}
(\byear{2012})
\doiurl{10.1007/s11069-012-0315-1}
\end{barticle}
\endbibitem

\bibitem[\protect\citeauthoryear{Sweet et~al.}{2018}]{Sweet2018}
\begin{botherref}
\oauthor{\bsnm{Sweet}, \binits{W.V.}},
\oauthor{\bsnm{Obeysekera}, \binits{J.T.B.}},
\oauthor{\bsnm{Marra}, \binits{J.J.}},
\oauthor{\bsnm{Dusek}, \binits{G.}}:
Patterns and projections of high tide flooding along the {U.S.} coastline using a common impact threshold.
(2018)
\doiurl{10.7289/V5/TR-NOS-COOPS-086}
\end{botherref}
\endbibitem

\bibitem[\protect\citeauthoryear{Feng et~al.}{2023}]{Feng2023}
\begin{barticle}
\bauthor{\bsnm{Feng}, \binits{J.}},
\bauthor{\bsnm{Li}, \binits{D.}},
\bauthor{\bsnm{Dang}, \binits{W.}},
\bauthor{\bsnm{Zhao}, \binits{L.}}:
\batitle{Changes in storm surges based on a bias-adjusted reconstruction dataset from 1900 to 2010}.
\bjtitle{Journal of Hydrology}
\bvolume{617},
\bfpage{128759}
(\byear{2023})
\end{barticle}
\endbibitem

\bibitem[\protect\citeauthoryear{Resio et~al.}{2017}]{Resio2017}
\begin{barticle}
\bauthor{\bsnm{Resio}, \binits{D.T.}},
\bauthor{\bsnm{J.~Powell}, \binits{N.}},
\bauthor{\bsnm{A.~Cialone}, \binits{M.}},
\bauthor{\bsnm{Das}, \binits{H.S.}},
\bauthor{\bsnm{Westerink}, \binits{J.J.}}:
\batitle{Quantifying impacts of forecast uncertainties on predicted storm surges}.
\bjtitle{Natural Hazards}
\bvolume{88}(\bissue{3}),
\bfpage{1423}--\blpage{1449}
(\byear{2017})
\doiurl{10.1007/s11069-017-2924-1}
\end{barticle}
\endbibitem

\bibitem[\protect\citeauthoryear{Butler et~al.}{2012}]{Butler2012}
\begin{barticle}
\bauthor{\bsnm{Butler}, \binits{T.}},
\bauthor{\bsnm{Altaf}, \binits{M.U.}},
\bauthor{\bsnm{Dawson}, \binits{C.}},
\bauthor{\bsnm{Hoteit}, \binits{I.}},
\bauthor{\bsnm{Luo}, \binits{X.}},
\bauthor{\bsnm{Mayo}, \binits{T.}}:
\batitle{Data assimilation within the {Advanced Circulation (ADCIRC)} modeling framework for hurricane storm surge forecasting}.
\bjtitle{Monthly Weather Review}
\bvolume{140}(\bissue{7}),
\bfpage{2215}--\blpage{2231}
(\byear{2012})
\doiurl{10.1175/mwr-d-11-00118.1}
\end{barticle}
\endbibitem

\bibitem[\protect\citeauthoryear{Muis et~al.}{2016}]{Muis2016}
\begin{botherref}
\oauthor{\bsnm{Muis}, \binits{S.}},
\oauthor{\bsnm{Verlaan}, \binits{M.}},
\oauthor{\bsnm{Winsemius}, \binits{H.C.}},
\oauthor{\bsnm{Aerts}, \binits{J.C.J.H.}},
\oauthor{\bsnm{Ward}, \binits{P.J.}}:
A global reanalysis of storm surges and extreme sea levels.
Nature Communications
\textbf{7}(1)
(2016)
\doiurl{10.1038/ncomms11969}
\end{botherref}
\endbibitem

\bibitem[\protect\citeauthoryear{Tadesse and Wahl}{2021}]{Tadesse2021}
\begin{botherref}
\oauthor{\bsnm{Tadesse}, \binits{M.G.}},
\oauthor{\bsnm{Wahl}, \binits{T.}}:
A database of global storm surge reconstructions.
Scientific Data
\textbf{8}(1)
(2021)
\doiurl{10.1038/s41597-021-00906-x}
\end{botherref}
\endbibitem

\bibitem[\protect\citeauthoryear{Kaiser et~al.}{2023}]{CERA2023}
\begin{botherref}
\oauthor{\bsnm{Kaiser}, \binits{C.}},
\oauthor{\bsnm{Dawson}, \binits{C.N.}},
\oauthor{\bsnm{Nikidis}, \binits{E.}},
\oauthor{\bsnm{Fleming}, \binits{J.G.}}:
{ADCIRC/SWAN} Hindcasts for Historical Storms 2003-2022.
Designsafe-CI
(2023).
\doiurl{10.17603/DS2-B5GH-CE94} .
\url{https://www.designsafe-ci.org/data/browser/public/designsafe.storage.published/PRJ-3932/#details-5508251847528869395-242ac117-0001-012}
\end{botherref}
\endbibitem

\bibitem[\protect\citeauthoryear{Haigh et~al.}{2022}]{Haigh2022}
\begin{barticle}
\bauthor{\bsnm{Haigh}, \binits{I.D.}},
\bauthor{\bsnm{Marcos}, \binits{M.}},
\bauthor{\bsnm{Talke}, \binits{S.A.}},
\bauthor{\bsnm{Woodworth}, \binits{P.L.}},
\bauthor{\bsnm{Hunter}, \binits{J.R.}},
\bauthor{\bsnm{Hague}, \binits{B.S.}},
\bauthor{\bsnm{Arns}, \binits{A.}},
\bauthor{\bsnm{Bradshaw}, \binits{E.}},
\bauthor{\bsnm{Thompson}, \binits{P.}}:
\batitle{<scp>gesla</scp> version 3: A major update to the global higher‐frequency sea‐level dataset}.
\bjtitle{Geoscience Data Journal}
\bvolume{10}(\bissue{3}),
\bfpage{293}--\blpage{314}
(\byear{2022})
\doiurl{10.1002/gdj3.174}
\end{barticle}
\endbibitem

\bibitem[\protect\citeauthoryear{Soci et~al.}{2024}]{Soci2024}
\begin{barticle}
\bauthor{\bsnm{Soci}, \binits{C.}},
\bauthor{\bsnm{Hersbach}, \binits{H.}},
\bauthor{\bsnm{Simmons}, \binits{A.}},
\bauthor{\bsnm{Poli}, \binits{P.}},
\bauthor{\bsnm{Bell}, \binits{B.}},
\bauthor{\bsnm{Berrisford}, \binits{P.}},
\bauthor{\bsnm{Horányi}, \binits{A.}},
\bauthor{\bsnm{Muñoz‐Sabater}, \binits{J.}},
\bauthor{\bsnm{Nicolas}, \binits{J.}},
\bauthor{\bsnm{Radu}, \binits{R.}},
\bauthor{\bsnm{Schepers}, \binits{D.}},
\bauthor{\bsnm{Villaume}, \binits{S.}},
\bauthor{\bsnm{Haimberger}, \binits{L.}},
\bauthor{\bsnm{Woollen}, \binits{J.}},
\bauthor{\bsnm{Buontempo}, \binits{C.}},
\bauthor{\bsnm{Thépaut}, \binits{J.}}:
\batitle{The {ERA5} global reanalysis from 1940 to 2022}.
\bjtitle{Quarterly Journal of the Royal Meteorological Society}
\bvolume{150}(\bissue{764}),
\bfpage{4014}--\blpage{4048}
(\byear{2024})
\doiurl{10.1002/qj.4803}
\end{barticle}
\endbibitem

\bibitem[\protect\citeauthoryear{Muis et~al.}{2020}]{Muis2020}
\begin{botherref}
\oauthor{\bsnm{Muis}, \binits{S.}},
\oauthor{\bsnm{Apecechea}, \binits{M.I.}},
\oauthor{\bsnm{Dullaart}, \binits{J.}},
\oauthor{\bsnm{Lima~Rego}, \binits{J.}},
\oauthor{\bsnm{Madsen}, \binits{K.S.}},
\oauthor{\bsnm{Su}, \binits{J.}},
\oauthor{\bsnm{Yan}, \binits{K.}},
\oauthor{\bsnm{Verlaan}, \binits{M.}}:
A high-resolution global dataset of extreme sea levels, tides, and storm surges, including future projections.
Frontiers in Marine Science
\textbf{7}
(2020)
\doiurl{10.3389/fmars.2020.00263}
\end{botherref}
\endbibitem

\bibitem[\protect\citeauthoryear{}{}]{noaaHURDATReanalysis}
\begin{botherref}
{H}{U}{R}{D}{A}{T} {R}e-analysis --- aoml.noaa.gov.
\url{https://www.aoml.noaa.gov/hrd/hurdat/Data_Storm.html}.
[Accessed 03-10-2025]
\end{botherref}
\endbibitem

\bibitem[\protect\citeauthoryear{}{2025}]{CERAarchive}
\begin{botherref}
{C}{E}{R}{A} - {H}istorical {S}torm {A}rchive --- historicalstorms.coastalrisk.live.
\url{https://historicalstorms.coastalrisk.live/}.
[Accessed 13-10-2025]
(2025)
\end{botherref}
\endbibitem

\bibitem[\protect\citeauthoryear{Lu et~al.}{2021}]{Lu2021}
\begin{barticle}
\bauthor{\bsnm{Lu}, \binits{X.}},
\bauthor{\bsnm{Yu}, \binits{H.}},
\bauthor{\bsnm{Ying}, \binits{M.}},
\bauthor{\bsnm{Zhao}, \binits{B.}},
\bauthor{\bsnm{Zhang}, \binits{S.}},
\bauthor{\bsnm{Lin}, \binits{L.}},
\bauthor{\bsnm{Bai}, \binits{L.}},
\bauthor{\bsnm{Wan}, \binits{R.}}:
\batitle{Western north pacific tropical cyclone database created by the china meteorological administration}.
\bjtitle{Advances in Atmospheric Sciences}
\bvolume{38}(\bissue{4}),
\bfpage{690}--\blpage{699}
(\byear{2021})
\doiurl{10.1007/s00376-020-0211-7}
\end{barticle}
\endbibitem

\bibitem[\protect\citeauthoryear{KITAMOTO et~al.}{2023}]{Kitamoto2023}
\begin{bchapter}
\bauthor{\bsnm{KITAMOTO}, \binits{A.}},
\bauthor{\bsnm{HWANG}, \binits{J.}},
\bauthor{\bsnm{VUILLOD}, \binits{B.}},
\bauthor{\bsnm{GAUTIER}, \binits{L.}},
\bauthor{\bsnm{TIAN}, \binits{Y.}},
\bauthor{\bsnm{CLANUWAT}, \binits{T.}}:
\bctitle{Digital typhoon: Long-term satellite image dataset for the spatio-temporal modeling of tropical cyclones}.
In: \bbtitle{NeurIPS 2023 Datasets and Benchmarks (Spotlight)}
(\byear{2023})
\end{bchapter}
\endbibitem

\bibitem[\protect\citeauthoryear{Zhao et~al.}{2025}]{Zhao2025}
\begin{botherref}
\oauthor{\bsnm{Zhao}, \binits{J.}},
\oauthor{\bsnm{Cerrone}, \binits{A.}},
\oauthor{\bsnm{Valseth}, \binits{E.}},
\oauthor{\bsnm{Westerink}, \binits{L.}},
\oauthor{\bsnm{Dawson}, \binits{C.}}:
Storm surge in color: Rgb-encoded physics-aware deep learning for storm surge forecasting.
arXiv preprint arXiv:2506.21743
(2025)
\end{botherref}
\endbibitem

\bibitem[\protect\citeauthoryear{Han et~al.}{2025}]{Han2025}
\begin{barticle}
\bauthor{\bsnm{Han}, \binits{L.}},
\bauthor{\bsnm{Lu}, \binits{W.}},
\bauthor{\bsnm{Dong}, \binits{C.}}:
\batitle{{XAI} helps in storm surge forecasts: A case study for the southeastern chinese coasts}.
\bjtitle{Journal of Marine Science and Engineering}
\bvolume{13}(\bissue{5}),
\bfpage{896}
(\byear{2025})
\doiurl{10.3390/jmse13050896}
\end{barticle}
\endbibitem

\bibitem[\protect\citeauthoryear{Saviz~Naeini et~al.}{2025}]{SavizNaeini2025}
\begin{barticle}
\bauthor{\bsnm{Saviz~Naeini}, \binits{S.}},
\bauthor{\bsnm{Snaiki}, \binits{R.}},
\bauthor{\bsnm{Wu}, \binits{T.}}:
\batitle{Advancing spatio-temporal storm surge prediction with hierarchical deep neural networks}.
\bjtitle{Natural Hazards}
\bvolume{121}(\bissue{14}),
\bfpage{16317}--\blpage{16344}
(\byear{2025})
\doiurl{10.1007/s11069-025-07428-4}
\end{barticle}
\endbibitem

\bibitem[\protect\citeauthoryear{Zhu et~al.}{2025}]{Zhu2025}
\begin{barticle}
\bauthor{\bsnm{Zhu}, \binits{Z.}},
\bauthor{\bsnm{Wang}, \binits{Z.}},
\bauthor{\bsnm{Dong}, \binits{C.}},
\bauthor{\bsnm{Yu}, \binits{M.}},
\bauthor{\bsnm{Xie}, \binits{H.}},
\bauthor{\bsnm{Cao}, \binits{X.}},
\bauthor{\bsnm{Han}, \binits{L.}},
\bauthor{\bsnm{Qi}, \binits{J.}}:
\batitle{Physics informed neural network modelling for storm surge forecasting — a case study in the {Bohai Sea}, {China}}.
\bjtitle{Coastal Engineering}
\bvolume{197},
\bfpage{104686}
(\byear{2025})
\doiurl{10.1016/j.coastaleng.2024.104686}
\end{barticle}
\endbibitem

\bibitem[\protect\citeauthoryear{Sreeraj et~al.}{2025}]{Sreeraj2025}
\begin{barticle}
\bauthor{\bsnm{Sreeraj}, \binits{P.}},
\bauthor{\bsnm{Swapna}, \binits{P.}},
\bauthor{\bsnm{Singh}, \binits{M.}},
\bauthor{\bsnm{Krishnan}, \binits{R.}}:
\batitle{Improved storm surge prediction and extreme sea level future projections in the indian ocean using deep learning}.
\bjtitle{Environmental Research Letters}
\bvolume{20}(\bissue{8}),
\bfpage{084058}
(\byear{2025})
\doiurl{10.1088/1748-9326/ade9e0}
\end{barticle}
\endbibitem

\bibitem[\protect\citeauthoryear{Huang et~al.}{2024}]{Huang2024}
\begin{barticle}
\bauthor{\bsnm{Huang}, \binits{S.}},
\bauthor{\bsnm{Nie}, \binits{H.}},
\bauthor{\bsnm{Jiao}, \binits{J.}},
\bauthor{\bsnm{Chen}, \binits{H.}},
\bauthor{\bsnm{Xie}, \binits{Z.}}:
\batitle{Tidal level prediction model based on {VMD-LSTM} neural network}.
\bjtitle{Water}
\bvolume{16}(\bissue{17}),
\bfpage{2452}
(\byear{2024})
\doiurl{10.3390/w16172452}
\end{barticle}
\endbibitem

\bibitem[\protect\citeauthoryear{Shi et~al.}{2024}]{Shi2024}
\begin{barticle}
\bauthor{\bsnm{Shi}, \binits{X.}},
\bauthor{\bsnm{Chen}, \binits{P.}},
\bauthor{\bsnm{Ye}, \binits{Z.}},
\bauthor{\bsnm{Zhang}, \binits{X.}},
\bauthor{\bsnm{Wang}, \binits{W.}}:
\batitle{Tide level prediction during typhoons based on variable topology in graph convolution recurrent neural networks}.
\bjtitle{Ocean Engineering}
\bvolume{312},
\bfpage{119228}
(\byear{2024})
\doiurl{10.1016/j.oceaneng.2024.119228}
\end{barticle}
\endbibitem

\bibitem[\protect\citeauthoryear{Pachev et~al.}{2023}]{Pachev2023}
\begin{barticle}
\bauthor{\bsnm{Pachev}, \binits{B.}},
\bauthor{\bsnm{Arora}, \binits{P.}},
\bauthor{\bsnm{del-Castillo-Negrete}, \binits{C.}},
\bauthor{\bsnm{Valseth}, \binits{E.}},
\bauthor{\bsnm{Dawson}, \binits{C.}}:
\batitle{A framework for flexible peak storm surge prediction}.
\bjtitle{Coastal Engineering}
\bvolume{186},
\bfpage{104406}
(\byear{2023})
\doiurl{10.1016/j.coastaleng.2023.104406}
\end{barticle}
\endbibitem

\bibitem[\protect\citeauthoryear{Dotse et~al.}{2023}]{Dotse2023}
\begin{barticle}
\bauthor{\bsnm{Dotse}, \binits{S.-Q.}},
\bauthor{\bsnm{Larbi}, \binits{I.}},
\bauthor{\bsnm{Limantol}, \binits{A.M.}},
\bauthor{\bsnm{De~Silva}, \binits{L.C.}}:
\batitle{A review of the application of hybrid machine learning models to improve rainfall prediction}.
\bjtitle{Modeling Earth Systems and Environment}
\bvolume{10}(\bissue{1}),
\bfpage{19}--\blpage{44}
(\byear{2023})
\doiurl{10.1007/s40808-023-01835-x}
\end{barticle}
\endbibitem

\bibitem[\protect\citeauthoryear{Wei et~al.}{2025}]{Wei2025}
\begin{botherref}
\oauthor{\bsnm{Wei}, \binits{C.}},
\oauthor{\bsnm{Zhao}, \binits{X.}},
\oauthor{\bsnm{Liu}, \binits{Y.}},
\oauthor{\bsnm{Yang}, \binits{P.}},
\oauthor{\bsnm{Zhou}, \binits{Z.}},
\oauthor{\bsnm{Chen}, \binits{Y.}}:
Bias analysis and correction of {ERA5} reanalysis in the context of tropical cyclones.
Journal of Geophysical Research: Atmospheres
\textbf{130}(2)
(2025)
\doiurl{10.1029/2024jd042737}
\end{botherref}
\endbibitem

\bibitem[\protect\citeauthoryear{Li et~al.}{2025}]{Li2025}
\begin{barticle}
\bauthor{\bsnm{Li}, \binits{R.}},
\bauthor{\bsnm{Guilloteau}, \binits{C.}},
\bauthor{\bsnm{Foufoula-Georgiou}, \binits{E.}}:
\batitle{Added value of environmental variables for satellite precipitation retrieval: A temporal coevolution perspective and a machine learning integration assessment}.
\bjtitle{Geophysical Research Letters}
\bvolume{52}(\bissue{11}),
\bfpage{2025}--\blpage{116048}
(\byear{2025})
\end{barticle}
\endbibitem

\bibitem[\protect\citeauthoryear{Cerrone et~al.}{2025}]{Cerrone2025}
\begin{barticle}
\bauthor{\bsnm{Cerrone}, \binits{A.R.}},
\bauthor{\bsnm{Westerink}, \binits{L.G.}},
\bauthor{\bsnm{Ling}, \binits{G.}},
\bauthor{\bsnm{Blakely}, \binits{C.P.}},
\bauthor{\bsnm{Wirasaet}, \binits{D.}},
\bauthor{\bsnm{Dawson}, \binits{C.}},
\bauthor{\bsnm{Westerink}, \binits{J.J.}}:
\batitle{Correcting physics-based global tide and storm water level forecasts with the temporal fusion transformer}.
\bjtitle{Ocean Modelling}
\bvolume{195},
\bfpage{102509}
(\byear{2025})
\doiurl{10.1016/j.ocemod.2025.102509}
\end{barticle}
\endbibitem

\bibitem[\protect\citeauthoryear{Carneiro-Barros et~al.}{2025}]{CarneiroBarros2025}
\begin{barticle}
\bauthor{\bsnm{Carneiro-Barros}, \binits{J.E.}},
\bauthor{\bsnm{Majidi}, \binits{A.G.}},
\bauthor{\bsnm{Plomaritis}, \binits{T.}},
\bauthor{\bsnm{Fazeres-Ferradosa}, \binits{T.}},
\bauthor{\bsnm{Rosa-Santos}, \binits{P.}},
\bauthor{\bsnm{Taveira-Pinto}, \binits{F.}}:
\batitle{Coastal flooding hazards in northern {Portugal}: A practical large-scale evaluation of total water levels and swash regimes}.
\bjtitle{Water}
\bvolume{17}(\bissue{10}),
\bfpage{1478}
(\byear{2025})
\doiurl{10.3390/w17101478}
\end{barticle}
\endbibitem

\bibitem[\protect\citeauthoryear{Giaremis et~al.}{2024}]{giaremis2024storm}
\begin{barticle}
\bauthor{\bsnm{Giaremis}, \binits{S.}},
\bauthor{\bsnm{Nader}, \binits{N.}},
\bauthor{\bsnm{Dawson}, \binits{C.}},
\bauthor{\bsnm{Kaiser}, \binits{C.}},
\bauthor{\bsnm{Nikidis}, \binits{E.}},
\bauthor{\bsnm{Kaiser}, \binits{H.}}:
\batitle{Storm surge modeling in the {AI} era: Using {LSTM}-based machine learning for enhancing forecasting accuracy}.
\bjtitle{Coastal Engineering}
\bvolume{191},
\bfpage{104532}
(\byear{2024})
\end{barticle}
\endbibitem

\bibitem[\protect\citeauthoryear{Tedesco et~al.}{2024}]{Tedesco2024}
\begin{barticle}
\bauthor{\bsnm{Tedesco}, \binits{P.}},
\bauthor{\bsnm{Rabault}, \binits{J.}},
\bauthor{\bsnm{Sætra}, \binits{M.L.}},
\bauthor{\bsnm{Kristensen}, \binits{N.M.}},
\bauthor{\bsnm{Aarnes}, \binits{O.J.}},
\bauthor{\bsnm{Breivik}, \binits{Ã.}},
\bauthor{\bsnm{Mauritzen}, \binits{C.}},
\bauthor{\bsnm{Sætra}, \binits{Ã.}}:
\batitle{Bias correction of operational storm surge forecasts using neural networks}.
\bjtitle{Ocean Modelling}
\bvolume{188},
\bfpage{102334}
(\byear{2024})
\doiurl{10.1016/j.ocemod.2024.102334}
\end{barticle}
\endbibitem

\bibitem[\protect\citeauthoryear{Liao et~al.}{2024}]{Liao2024}
\begin{barticle}
\bauthor{\bsnm{Liao}, \binits{J.}},
\bauthor{\bsnm{Li}, \binits{Y.}},
\bauthor{\bsnm{Li}, \binits{J.}},
\bauthor{\bsnm{Li}, \binits{S.}},
\bauthor{\bsnm{Peng}, \binits{S.}}:
\batitle{A two-module bias-correction model for sea wave hindcasting based on the long-short term memory neural network}.
\bjtitle{Ocean Engineering}
\bvolume{311},
\bfpage{118827}
(\byear{2024})
\doiurl{10.1016/j.oceaneng.2024.118827}
\end{barticle}
\endbibitem

\bibitem[\protect\citeauthoryear{Zhang et~al.}{2024a}]{Zhang2024}
\begin{botherref}
\oauthor{\bsnm{Zhang}, \binits{S.}},
\oauthor{\bsnm{Harrop}, \binits{B.}},
\oauthor{\bsnm{Leung}, \binits{L.R.}},
\oauthor{\bsnm{Charalampopoulos}, \binits{A.}},
\oauthor{\bsnm{Barthel~Sorensen}, \binits{B.}},
\oauthor{\bsnm{Xu}, \binits{W.}},
\oauthor{\bsnm{Sapsis}, \binits{T.}}:
A machine learning bias correction on large‐scale environment of high‐impact weather systems in {E3SM} atmosphere model.
Journal of Advances in Modeling Earth Systems
\textbf{16}(8)
(2024)
\doiurl{10.1029/2023ms004138}
\end{botherref}
\endbibitem

\bibitem[\protect\citeauthoryear{Zhang et~al.}{2024b}]{Zhang2024b}
\begin{barticle}
\bauthor{\bsnm{Zhang}, \binits{W.}},
\bauthor{\bsnm{Sun}, \binits{Y.}},
\bauthor{\bsnm{Wu}, \binits{Y.}},
\bauthor{\bsnm{Dong}, \binits{J.}},
\bauthor{\bsnm{Song}, \binits{X.}},
\bauthor{\bsnm{Gao}, \binits{Z.}},
\bauthor{\bsnm{Pang}, \binits{R.}},
\bauthor{\bsnm{Guoan}, \binits{B.}}:
\batitle{A deep-learning real-time bias correction method for significant wave height forecasts in the {Western North Pacific}}.
\bjtitle{Ocean Modelling}
\bvolume{187},
\bfpage{102289}
(\byear{2024})
\doiurl{10.1016/j.ocemod.2023.102289}
\end{barticle}
\endbibitem

\bibitem[\protect\citeauthoryear{Kao et~al.}{2024}]{Kao2024}
\begin{barticle}
\bauthor{\bsnm{Kao}, \binits{Y.-C.}},
\bauthor{\bsnm{Tsou}, \binits{H.-E.}},
\bauthor{\bsnm{Chen}, \binits{C.-J.}}:
\batitle{Development of multi-source weighted-ensemble precipitation: Influence of bias correction based on recurrent convolutional neural networks}.
\bjtitle{Journal of Hydrology}
\bvolume{629},
\bfpage{130621}
(\byear{2024})
\doiurl{10.1016/j.jhydrol.2024.130621}
\end{barticle}
\endbibitem

\bibitem[\protect\citeauthoryear{Liu et~al.}{2023}]{Liu2023}
\begin{botherref}
\oauthor{\bsnm{Liu}, \binits{G.}},
\oauthor{\bsnm{Bracco}, \binits{A.}},
\oauthor{\bsnm{Brajard}, \binits{J.}}:
Systematic bias correction in ocean mesoscale forecasting using machine learning.
Journal of Advances in Modeling Earth Systems
\textbf{15}(11)
(2023)
\doiurl{10.1029/2022ms003426}
\end{botherref}
\endbibitem

\bibitem[\protect\citeauthoryear{Yan et~al.}{2018}]{Yan2018}
\begin{botherref}
\oauthor{\bsnm{Yan}, \binits{S.}},
\oauthor{\bsnm{Xiong}, \binits{Y.}},
\oauthor{\bsnm{Lin}, \binits{D.}}:
Spatial Temporal Graph Convolutional Networks for Skeleton-Based Action Recognition.
arXiv
(2018).
\doiurl{10.48550/ARXIV.1801.07455} .
\url{https://arxiv.org/abs/1801.07455}
\end{botherref}
\endbibitem

\bibitem[\protect\citeauthoryear{Zhang et~al.}{2019}]{Zhang2019b}
\begin{barticle}
\bauthor{\bsnm{Zhang}, \binits{C.}},
\bauthor{\bsnm{Yu}, \binits{J.J.Q.}},
\bauthor{\bsnm{Liu}, \binits{Y.}}:
\batitle{Spatial-temporal graph attention networks: A deep learning approach for traffic forecasting}.
\bjtitle{IEEE Access}
\bvolume{7},
\bfpage{166246}--\blpage{166256}
(\byear{2019})
\doiurl{10.1109/access.2019.2953888}
\end{barticle}
\endbibitem

\bibitem[\protect\citeauthoryear{Jiang et~al.}{2024}]{Jiang2024}
\begin{barticle}
\bauthor{\bsnm{Jiang}, \binits{W.}},
\bauthor{\bsnm{Zhang}, \binits{J.}},
\bauthor{\bsnm{Li}, \binits{Y.}},
\bauthor{\bsnm{Zhang}, \binits{D.}},
\bauthor{\bsnm{Hu}, \binits{G.}},
\bauthor{\bsnm{Gao}, \binits{H.}},
\bauthor{\bsnm{Duan}, \binits{Z.}}:
\batitle{Advancing storm surge forecasting from scarce observation data: A causal-inference based spatio-temporal graph neural network approach}.
\bjtitle{Coastal Engineering}
\bvolume{190},
\bfpage{104512}
(\byear{2024})
\doiurl{10.1016/j.coastaleng.2024.104512}
\end{barticle}
\endbibitem

\bibitem[\protect\citeauthoryear{Kazadi et~al.}{2024}]{Kazadi2024}
\begin{botherref}
\oauthor{\bsnm{Kazadi}, \binits{A.}},
\oauthor{\bsnm{Doss-Gollin}, \binits{J.}},
\oauthor{\bsnm{Sebastian}, \binits{A.}},
\oauthor{\bsnm{Silva}, \binits{A.}}:
{FloodGNN}-{GRU}: a spatio-temporal graph neural network for flood prediction.
Environmental Data Science
\textbf{3}
(2024)
\doiurl{10.1017/eds.2024.19}
\end{botherref}
\endbibitem

\bibitem[\protect\citeauthoryear{Lam et~al.}{2023}]{Lam2023}
\begin{barticle}
\bauthor{\bsnm{Lam}, \binits{R.}},
\bauthor{\bsnm{Sanchez-Gonzalez}, \binits{A.}},
\bauthor{\bsnm{Willson}, \binits{M.}},
\bauthor{\bsnm{Wirnsberger}, \binits{P.}},
\bauthor{\bsnm{Fortunato}, \binits{M.}},
\bauthor{\bsnm{Alet}, \binits{F.}},
\bauthor{\bsnm{Ravuri}, \binits{S.}},
\bauthor{\bsnm{Ewalds}, \binits{T.}},
\bauthor{\bsnm{Eaton-Rosen}, \binits{Z.}},
\bauthor{\bsnm{Hu}, \binits{W.}},
\bauthor{\bsnm{Merose}, \binits{A.}},
\bauthor{\bsnm{Hoyer}, \binits{S.}},
\bauthor{\bsnm{Holland}, \binits{G.}},
\bauthor{\bsnm{Vinyals}, \binits{O.}},
\bauthor{\bsnm{Stott}, \binits{J.}},
\bauthor{\bsnm{Pritzel}, \binits{A.}},
\bauthor{\bsnm{Mohamed}, \binits{S.}},
\bauthor{\bsnm{Battaglia}, \binits{P.}}:
\batitle{Learning skillful medium-range global weather forecasting}.
\bjtitle{Science}
\bvolume{382}(\bissue{6677}),
\bfpage{1416}--\blpage{1421}
(\byear{2023})
\doiurl{10.1126/science.adi2336}
\end{barticle}
\endbibitem

\bibitem[\protect\citeauthoryear{Wu et~al.}{2023}]{Wu2023}
\begin{botherref}
\oauthor{\bsnm{Wu}, \binits{B.}},
\oauthor{\bsnm{Chen}, \binits{W.}},
\oauthor{\bsnm{Wang}, \binits{W.}},
\oauthor{\bsnm{Peng}, \binits{B.}},
\oauthor{\bsnm{Sun}, \binits{L.}},
\oauthor{\bsnm{Chen}, \binits{L.}}:
WeatherGNN: Exploiting Meteo- and Spatial-Dependencies for Local Numerical Weather Prediction Bias-Correction.
arXiv
(2023).
\doiurl{10.48550/ARXIV.2310.05517} .
\url{https://arxiv.org/abs/2310.05517}
\end{botherref}
\endbibitem

\bibitem[\protect\citeauthoryear{Nader et~al.}{2015}]{nader2015classification}
\begin{bchapter}
\bauthor{\bsnm{Nader}, \binits{N.}},
\bauthor{\bsnm{Hassan}, \binits{M.}},
\bauthor{\bsnm{Falou}, \binits{W.}},
\bauthor{\bsnm{Diab}, \binits{A.}},
\bauthor{\bsnm{Al-Omar}, \binits{S.}},
\bauthor{\bsnm{Khalil}, \binits{M.}},
\bauthor{\bsnm{Marque}, \binits{C.}}:
\bctitle{Classification of pregnancy and labor contractions using a graph theory based analysis}.
In: \bbtitle{2015 37th Annual International Conference of the IEEE Engineering in Medicine and Biology Society (EMBC)},
pp. \bfpage{2876}--\blpage{2879}
(\byear{2015}).
\bcomment{IEEE}
\end{bchapter}
\endbibitem

\bibitem[\protect\citeauthoryear{Nader et~al.}{2016}]{nader2016node}
\begin{bchapter}
\bauthor{\bsnm{Nader}, \binits{N.}},
\bauthor{\bsnm{Hassan}, \binits{M.}},
\bauthor{\bsnm{Falou}, \binits{W.}},
\bauthor{\bsnm{Marque}, \binits{C.}},
\bauthor{\bsnm{Khalil}, \binits{M.}}:
\bctitle{A node-wise analysis of the uterine muscle networks for pregnancy monitoring}.
In: \bbtitle{2016 38th Annual International Conference of the IEEE Engineering in Medicine and Biology Society (EMBC)},
pp. \bfpage{712}--\blpage{715}
(\byear{2016}).
\bcomment{IEEE}
\end{bchapter}
\endbibitem

\bibitem[\protect\citeauthoryear{Al-Omar et~al.}{2015}]{al2015detecting}
\begin{bchapter}
\bauthor{\bsnm{Al-Omar}, \binits{S.}},
\bauthor{\bsnm{Diab}, \binits{A.}},
\bauthor{\bsnm{Nader}, \binits{N.}},
\bauthor{\bsnm{Khalil}, \binits{M.}},
\bauthor{\bsnm{Karlsson}, \binits{B.}},
\bauthor{\bsnm{Marque}, \binits{C.}}:
\bctitle{Detecting labor using graph theory on connectivity matrices of uterine emg}.
In: \bbtitle{2015 37th Annual International Conference of the IEEE Engineering in Medicine and Biology Society (EMBC)},
pp. \bfpage{2195}--\blpage{2198}
(\byear{2015}).
\bcomment{IEEE}
\end{bchapter}
\endbibitem

\bibitem[\protect\citeauthoryear{Pedregosa et~al.}{2011}]{pedregosa2011scikit}
\begin{barticle}
\bauthor{\bsnm{Pedregosa}, \binits{F.}},
\bauthor{\bsnm{Varoquaux}, \binits{G.}},
\bauthor{\bsnm{Gramfort}, \binits{A.}},
\bauthor{\bsnm{Michel}, \binits{V.}},
\bauthor{\bsnm{Thirion}, \binits{B.}},
\bauthor{\bsnm{Grisel}, \binits{O.}},
\bauthor{\bsnm{Blondel}, \binits{M.}},
\bauthor{\bsnm{Prettenhofer}, \binits{P.}},
\bauthor{\bsnm{Weiss}, \binits{R.}},
\bauthor{\bsnm{Dubourg}, \binits{V.}}, \betal:
\batitle{Scikit-learn: Machine learning in python}.
\bjtitle{the Journal of machine Learning research}
\bvolume{12},
\bfpage{2825}--\blpage{2830}
(\byear{2011})
\end{barticle}
\endbibitem

\bibitem[\protect\citeauthoryear{G{\'e}ron}{2022}]{geron2022hands}
\begin{bbook}
\bauthor{\bsnm{G{\'e}ron}, \binits{A.}}:
\bbtitle{Hands-on Machine Learning with Scikit-Learn, Keras, and TensorFlow}.
\bpublisher{" O'Reilly Media, Inc."}, \blocation{???}
(\byear{2022})
\end{bbook}
\endbibitem

\bibitem[\protect\citeauthoryear{}{}]{SaffirSimpson}
\begin{botherref}
{S}affir-{S}impson {H}urricane {S}cale --- weather.gov.
\url{https://www.weather.gov/mfl/saffirsimpson}
\end{botherref}
\endbibitem

\bibitem[\protect\citeauthoryear{Vaswani et~al.}{2017}]{vaswani2017attention}
\begin{botherref}
\oauthor{\bsnm{Vaswani}, \binits{A.}},
\oauthor{\bsnm{Shazeer}, \binits{N.}},
\oauthor{\bsnm{Parmar}, \binits{N.}},
\oauthor{\bsnm{Uszkoreit}, \binits{J.}},
\oauthor{\bsnm{Jones}, \binits{L.}},
\oauthor{\bsnm{Gomez}, \binits{A.N.}},
\oauthor{\bsnm{Kaiser}, \binits{{\L}.}},
\oauthor{\bsnm{Polosukhin}, \binits{I.}}:
Attention is all you need.
Advances in neural information processing systems
\textbf{30}
(2017)
\end{botherref}
\endbibitem

\bibitem[\protect\citeauthoryear{Kingma and Ba}{2014}]{Kingma2014}
\begin{botherref}
\oauthor{\bsnm{Kingma}, \binits{D.P.}},
\oauthor{\bsnm{Ba}, \binits{J.}}:
Adam: A Method for Stochastic Optimization.
arXiv
(2014).
\doiurl{10.48550/ARXIV.1412.6980}
\end{botherref}
\endbibitem

\bibitem[\protect\citeauthoryear{Cangialosi and Alaka}{2023}]{Cangialosi2023}
\begin{botherref}
\oauthor{\bsnm{Cangialosi}, \binits{J.P.}},
\oauthor{\bsnm{Alaka}, \binits{L.}}:
National Hurricane Center Tropical Cyclone Report - Hurricane Idalia (AL102023), 26–31 August 2023.
\url{https://www.nhc.noaa.gov/data/tcr/AL102023_Idalia.pdf}.
National Hurricane Center Tropical Cyclone Report
(2023)
\end{botherref}
\endbibitem

\bibitem[\protect\citeauthoryear{{NHC - National Hurricane Center and Central Pacific National Center}}{2025}]{NHC}
\begin{botherref}
\oauthor{\bsnm{{NHC - National Hurricane Center and Central Pacific National Center}}}:
{NHC} Active Tropical Cyclones.
\url{https://www.nhc.noaa.gov/cyclones/}
(2025)
\end{botherref}
\endbibitem

\bibitem[\protect\citeauthoryear{}{}]{NOAAInundationHistory}
\begin{botherref}
{I}nundation {H}istory - {N}{O}{A}{A} {T}ides \& {C}urrents --- tidesandcurrents.noaa.gov.
\url{https://tidesandcurrents.noaa.gov/inundationdb/inundation.html}.
[Accessed 15-10-2025]
\end{botherref}
\endbibitem

\bibitem[\protect\citeauthoryear{{D}epartment~of {C}ommerce}{2025}]{NOAA2017}
\begin{botherref}
\oauthor{\bsnm{{C}ommerce}, \binits{N.W.S.} \bsuffix{{N}ational {O}ceanic \& {A}tmospheric~{A}dministration}}:
{N}ational {W}eather {S}ervice {I}nstruction 10-320. {S}urf {Z}one {F}orecast and {C}oastal/{L}akeshore {H}azard {S}ervices.
\url{https://www.weather.gov/media/directives/010_pdfs/pd01003020curr.pdf}.
[Accessed 16-10-2025]
(2025)
\end{botherref}
\endbibitem

\end{thebibliography}



\end{document}